\documentclass[pdflatex,sn-nature]{sn-jnl}% Style for submissions to Nature Portfolio journals
\usepackage{graphicx}%
\usepackage{multirow}%
\usepackage{amsmath,amssymb,amsfonts}%
\usepackage{amsthm}%
\usepackage{mathrsfs}%
\usepackage[title]{appendix}%
\usepackage[table]{xcolor}
\usepackage{textcomp}%
\usepackage{manyfoot}%
\usepackage{booktabs}%
\usepackage{algorithm}%
\usepackage{algorithmicx}%
\usepackage{algpseudocode}%
\usepackage{listings}%
\usepackage{makecell}
\usepackage{caption}       % 标题格式控制
\usepackage{svg}
\usepackage{graphicx}
\usepackage{adjustbox}
\usepackage{tabularx}
\usepackage{colortbl}
\usepackage{url}
\usepackage{verbatim}
\usepackage{array}
\usepackage[T1]{fontenc}
\usepackage[utf8]{inputenc}

\theoremstyle{thmstyleone}%
\theoremstyle{thmstyletwo}%

\theoremstyle{thmstylethree}%

\begin{document}

\title[Article Title]{Model Hemorrhage and the Robustness Limits of Large Language Models}


\author[1]{\fnm{Ziyang} \sur{Ma}}\email{maziyang@whu.edu.cn}

\author*[2]{\fnm{Zuchao} \sur{Li}}\email{zcli-charlie@whu.edu.cn}

\author[1]{\fnm{Lefei} \sur{Zhang}}\email{zhanglefei@whu.edu.cn}

\author[2]{\fnm{Gui-Song} \sur{Xia}}\email{guisong.xia@whu.edu.cn}

\author[1]{\fnm{Bo} \sur{Du}}\email{dubo@whu.edu.cn}

\author[3]{\fnm{Liangpei} \sur{Zhang}}\email{zlp62@whu.edu.cn}

\author[4]{\fnm{Dacheng} \sur{Tao}}\email{dacheng.tao@gmail.com}

\affil[1]{\orgdiv{School of Computer Science}, \orgname{Wuhan University}, \orgaddress{\city{Wuhan}, \country{China}}}

\affil*[2]{\orgdiv{School of Artificial Intelligence}, \orgname{Wuhan University}, \orgaddress{\city{Wuhan}, \country{China}}}

\affil[3]{\orgdiv{State Key Laboratory of Information Engineering in Surveying, Mapping and Remote Sensing}, \orgname{{Wuhan University}, \orgaddress\city{Wuhan}, \country{China}}}

\affil[4]{\orgdiv{Nanyang Technological University}, \orgname{School of Computer Science and Engineering}, \orgaddress{\city{Singapore},  \country{Singapore}}}

%\author*[1,2]{\fnm{First} \sur{Author}}\email{iauthor@gmail.com}



%\affil*[1]{\orgdiv{Department}, \orgname{Organization}, %\orgaddress{\street{Street}, \city{City}, \postcode{100190}, \state{State}, \country{Country}}}

%\affil[2]{\orgdiv{Department}, \orgname{Organization}, \orgaddress{\street{Street}, \city{City}, \postcode{10587}, \state{State}, \country{Country}}}

%\affil[3]{\orgdiv{Department}, \orgname{Organization}, \orgaddress{\street{Street}, \city{City}, \postcode{610101}, \state{State}, \country{Country}}}


\abstract{Large language models (LLMs) have demonstrated exceptional performance across a broad spectrum of natural language processing tasks. However, as these models undergo various modifications for deployment, such as quantization, pruning, and decoding strategy adjustments, they are prone to significant performance degradation. We term this phenomenon “Model Hemorrhage,” referring to the decline in performance and accuracy caused by parameter alterations and architectural modifications. This paper introduces the concept of model hemorrhage to systematically examine the circumstances under which LLMs experience such degradation and the inherent robustness of different frameworks and model types. Through extensive analysis, we identify specific operations—such as layer expansion, compression techniques, and decoding adjustments—that frequently lead to model hemorrhage, as well as the underlying causes. Additionally, we propose potential strategies for mitigating these issues, aiming to enhance the stability, reliability, and scalability of LLMs in diverse application contexts. This work is a foundational exploration into model robustness, guiding further research on preserving model integrity amid modification-induced challenges.}
 



\maketitle

\section{Introduction}\label{sec1}

Large language models (LLMs) have rapidly advanced since the introduction of the Transformer architecture. Built on the core concept of "Attention is All You Need" \cite{ref1}, this innovation has fundamentally reshaped the field of natural language processing (NLP). Unlike traditional Recurrent Neural Networks (RNNs) and Long Short-Term Memory (LSTM) networks, the multi-head attention mechanism in Transformers efficiently handles long-range dependencies while greatly improving parallelization capabilities. This feature allows models to scale up, capturing more complex semantic relationships in language. Transformers have laid the groundwork for a series of groundbreaking models, with BERT \cite{ref8} and GPT \cite{ref5} being the most prominent examples. Their success marks a transition from the encoder-decoder paradigm.

In recent years, as computational power has improved and data resources have accumulated, the scale of LLMs has grown exponentially. The number of parameters has surged from hundreds of millions to hundreds of billions and even trillions, leading to substantial improvements in model performance. This phenomenon is described by the "Scaling Law," \cite{kaplan2020scaling} which suggests that model performance continues to enhance as parameters, training data, and computational resources increase. However, along with the growth in model size, computational costs and training time have also escalated significantly. To address this, the concept of "Compute-Optimal Large Language Models"  \cite{ref13} was introduced, prompting researchers to explore the complex relationships between model quality, data volume, model size, and computation time.

As LLMs scale to larger sizes, traditional single-model architectures face dual challenges in performance and efficiency. Sparse Mixture of Experts (MoE) offers a promising solution, enabling effective scaling of model capacity while maintaining fixed computational costs. The core idea behind MoE is to allocate tasks to different expert networks through an appropriate gating strategy, reducing computational overhead while maintaining or even improving the model’s expressive capability and performance. Recently, the launch of the open-source MoE model DeepSeek-R1 \cite{deepseek_r1} has provided new insights into LLM research. DeepSeek-R1 integrates advanced MoE gating mechanisms and reinforcement learning strategies, driving a paradigm shift from pretraining to inference computation. This innovation not only optimizes computational efficiency but also offers effective solutions to the task diversity and computational limitations encountered in practical applications, opening new avenues for the future development of LLMs.

As model scale increases, large language models (LLMs) have progressed from single-task applications to multi-task capabilities and from unimodal functionality to multimodal integration, paving the way toward the realization of a "world model." This evolution represents a significant milestone in natural language processing (NLP) and unleashes unprecedented potential for artificial intelligence (AI) across diverse domains.

Initially, LLMs were applied to single tasks such as text generation and classification. Today, their capabilities encompass multimodal tasks, including image captioning, video analysis, and speech recognition, with seamless transitions across modalities to provide cross-domain solutions. In multi-task processing, the GPT series exemplifies remarkable versatility, excelling in text generation while performing machine translation, document summarization, and context-aware conversational tasks within a unified framework \cite{ref4, ref5}.

In multimodality, LLMs have expanded their frontiers, from early text-to-image models like DALL-E \cite{ref6} to image-text retrieval systems such as CLIP \cite{radford2021learning}. These models not only process text but also integrate images and audio, enabling robust solutions for real-world challenges. By consolidating multimodal data within a unified architecture, they mark a critical step toward realizing general artificial intelligence (General AI).

In this Perspective, we propose a novel theoretical framework named ``Model Hemorrhage'' to systematically and comprehensively address the performance degradation issues faced by large-scale language models during optimization and deployment processes. The primary objectives of this survey are as follows:

\begin{enumerate}
    \item Provide a clear and systematic definition of ``Model Hemorrhage'' to expand the boundaries of traditional robustness research in large-scale models.
    
    \item Review recent advancements in model compression, optimization, parameter scaling, and architectural innovations, while discussing the causes and phenomena of accuracy loss and performance degradation during model optimization and deployment. This establishes the theoretical foundation for the ``Model Hemorrhage'' framework.
    
    \item Conduct systematic experiments on common deployment operations, including pruning, quantization, and inference optimization algorithms, to investigate model sensitivity and validate the feasibility of the theoretical framework.
    
    \item Provide a detailed discussion on the limitations of current robustness research in large-scale models, challenges in model deployment, and promising future directions to address the evolving challenges of large language models in dynamic environments.
\end{enumerate}

The remainder of this survey is organized as follows: In Section~\ref{sec:definition}, we introduce the definition of ``Model Hemorrhage'' through observed degradation phenomena in large-scale model deployment. Section~\ref{sec:analysis} comprehensively analyzes the relationship between robustness and critical factors such as compression deployment, inference methods, parameter scaling, and model architectures.  Section~\ref{sec:challenges} discusses challenges and future research directions.
Section~\ref{sec:conclusion} concludes with the significance and implications of the ``Model Hemorrhage'' framework.
Appendix~\ref{sec:experiments} presents experimental investigations into the sensitivity of large models to common Model Hemorrhage operations.

\section{Challenges in Scaling and Deploying LLMs}\label{sec:definition}
\subsection{Key Challenges in LLM Scaling and Deployment}
The rapid scaling of large language models (LLMs) has ushered in groundbreaking advancements in natural language processing (NLP), but it has also introduced significant challenges in their training, deployment, and real-world applications. These challenges are particularly evident in the following aspects:
\subsubsection{Performance-Computation Trade-offs} 
As large language models (LLMs) continue to grow in size, deploying them often requires architectural optimizations and parameter adjustments to accommodate computational resource constraints. While these adjustments effectively reduce inference costs, they also introduce significant performance degradation.

With increasingly larger models and datasets, the computational demands for training and inference have risen sharply. For instance, training Google's PaLM, a large-scale language model with 540 billion parameters, required an estimated 8.4 million TPU hours \cite{ref26}. This computational cost represents a significant barrier to entry, affordable only to organizations with access to specialized hardware and extensive computational resources. During practical deployment, memory consumption for storing activation parameters represents only a fraction of the total memory requirements. A key challenge in inference is the memory limitation imposed by autoregressive models, particularly in handling key-value caches (KV caches), which are a critical area of ongoing research. For a comprehensive review of methods to optimize KV cache consumption, see \cite{luohekeep}.
The total memory required during inference, \(M_{\text{total}}\), can be approximated by the formula:
\begin{equation}
M_{\text{total}} = B \cdot S \cdot \left( L \cdot H \cdot P \cdot 2 \right) + M_{\text{activation}} + M_{\text{gradient}}
\end{equation}
where \(B\) is the batch size, \(S\) is the sequence length, \(L\) is the number of layers, \(H\) is the hidden size, and \(P\) represents the numerical precision (e.g., FP32 = 4 bytes, FP16 = 2 bytes). \(M_{\text{activation}}\) refers to the memory required to store activations, while \(M_{\text{gradient}}\) represents the memory required for gradient storage, which is typically zero during inference but becomes significant during training. For example, in the case of OPT-30B, assuming half-precision inference (\(P = 2\) bytes), a maximum sequence length of 1024, and a batch size of 128, the total KV cache requirement amounts to 180GB, whereas the model parameters alone occupy only \(2 \cdot 30B = 60GB\). 

Consequently, deploying LLMs in practical applications often necessitates architectural modifications such as quantization, pruning, or sparsity techniques. Although these optimization methods reduce memory and computational costs, they are frequently accompanied by performance degradation, particularly in terms of accuracy, robustness, and generalization capabilities.

\subsubsection{Costs of Model Scaling} 

The scaling of large language models (LLMs) has significantly advanced the field of natural language processing (NLP), with parameter sizes growing from millions to hundreds of billions, and even trillions. This expansion has enabled greater task adaptability and improved generative quality. However, scaling also introduces several costs and challenges, particularly in terms of model redundancy, diminishing marginal returns, degradation phenomena, and bottlenecks in resource consumption and efficiency.\\
\textbf{Computational Costs.} Under the guidance of scaling laws, the increase in model parameters is accompanied by the growth of training datasets and the extension of training steps, leading to a dramatic rise in resource consumption for training and inference. For instance, the carbon emissions from training GPT-3 are estimated at approximately 552 \text{tCO\textsubscript{2}e}
, which is equivalent to the emissions produced by a typical gasoline-powered car driving around the Earth's equator nearly 69 times \cite{ref27}.\\
\textbf{Diminishing Marginal Returns.} As parameter sizes grow, the performance improvements of LLMs follow a law of diminishing returns. \cite{kaplan2020scaling} observed that when model parameters exceed hundreds of billions, additional parameters contribute only marginal gains to performance. Furthermore, parameter expansion exacerbates gradient optimization challenges, particularly in excessively deep or wide networks, where issues such as vanishing or exploding gradients become more prominent. This can significantly slow training convergence or even lead to training failure.\\
\textbf{Structural Redundancy and Degradation.} Recent studies highlight significant structural redundancy in large-scale models, which will be discussed in detail in \ref{sec:Redundancy}. Such redundancy results in resource inefficiency and negatively impacts task-specific performance. For instance, excessive parameters can hinder adaptability in low-resource or small-sample tasks. In short-text generation, this may manifest as overly verbose or irrelevant outputs. These models also exhibit overfitting tendencies and unstable parameter optimization. Additionally, while increased parameter counts enhance model complexity, they can make models more sensitive to adversarial samples and input noise, further reducing interpretability. These issues are particularly critical in high-stakes domains such as healthcare and law, where reliability and explainability are paramount.\\

\subsubsection{Decoding Strategy Limitations}
In the practical deployment of large language models (LLMs), decoding strategies often represent a trade-off between quality and efficiency. These strategies directly influence the quality and applicability of the model's generated content. However, due to the complexity of the generation process and resource constraints, adjustments to decoding strategies frequently lead to significant degradation in generation quality.

Greedy decoding, while highly efficient, tends to produce repetitive and verbose outputs, particularly in long-text generation tasks. Random sampling, especially with higher temperature values, can result in semantically incoherent or logically inconsistent content. Beam search, although improving generation quality, often sacrifices content diversity. Moreover, decoding strategies with higher complexity may lead to slower decoding speeds, further complicating their deployment in resource-constrained environments.

\subsubsection{Multimodal Model Challenges} 
Multimodal models show great potential in handling complex tasks, but they often face challenges such as modality inconsistency, data noise, missing information, or modality bias in practical applications.\\
\textbf{Data Noise.} Data noise is a critical factor affecting model performance. Real-world data often contains modality-specific noise. For example, images may suffer from distortions, variations in brightness, stylization, or weather conditions. Text data can include OCR errors, synonyms, rare words, or shifts in tense and voice. Such noise interferes with learning cross-modal associations, leading to instability in task performance.\\
\textbf{Modality Misalignment.} Misalignment between modalities significantly increases the complexity of multimodal tasks. In many scenarios, temporal or semantic alignment across modalities is imprecise. For instance, textual descriptions may deviate from image content. This inconsistency makes it difficult for models to effectively learn cross-modal correlations, directly reducing prediction accuracy.\\
\textbf{Missing Modalities.} Missing modality data is a common issue in practical applications. In dynamic environments or resource-constrained settings, some modalities may be unavailable due to limitations in data collection or operational constraints. For example, in medical diagnostics, certain imaging data may be unavailable due to equipment or privacy restrictions. This requires models to handle incomplete modality inputs effectively.\\
\textbf{Modality Imbalance.} Modality imbalance is a core challenge in multimodal model design. Some modalities may contribute the majority of information, while others play a minor role. For instance, in video analysis, visual data often dominates, whereas audio information may be less significant. This imbalance can lead to over-reliance on dominant modalities during training, limiting the effective utilization of auxiliary modalities.

\subsection{Defining Model Hemorrhage}
\textbf{Model Hemorrhage} refers to the phenomenon where large language models (LLMs) and their extended frameworks (e.g., multimodal models) experience performance degradation, robustness weakening, or adaptability failure during training, optimization, deployment, or task adaptation. This phenomenon is attributed to intrinsic model characteristics, external architectural adjustments, parameter changes, scaling, or data complexities. Manifestations include, but are not limited to, reduced generation quality, unstable task performance, increased vulnerability to adversarial samples, constrained generalization ability, difficulties in multimodal task adaptation, and heightened sensitivity to input noise or modality misalignment. The core characteristics of model hemorrhage include:
\begin{enumerate}
\item Performance Degradation. The performance of models can decline significantly across various tasks, including classification, translation, or generation, particularly in low-resource or distribution-shift scenarios. Generated content may become verbose, repetitive, or lack logical and semantic coherence. For example, in short-text generation, models frequently produce redundant or irrelevant content, undermining output quality.
    
\item Robustness Weakening. Models exhibit increased sensitivity to minor perturbations, such as adversarial samples, causing substantial deviations in output. Additionally, noise in input data, such as OCR errors or blurry images, destabilizes task execution. This issue is particularly pronounced in multimodal tasks, where noise disrupts cross-modal learning and representation consistency.

\item Generalization Limitations. Models often struggle to adapt effectively to tasks with distributions different from the training data, leading to degraded cross-domain performance. In multimodal tasks, over-reliance on dominant modalities reduces the contributions of auxiliary modalities. This imbalance negatively impacts the overall performance and generalizability of the model.

\item Resource Efficiency Challenges. Inference performance deteriorates due to architectural optimizations like quantization or pruning, which, while reducing computational costs, compromise the model’s representational capabilities. Furthermore, high computational demands and latency issues in resource-constrained environments hinder real-time applications, as models require substantial memory and processing power for inference.
    
\end{enumerate}

\section{Mechanisms and Triggers of Model Hemorrhage}\label{sec:analysis}
\begin{figure*}[ht]
    \centering
    \includegraphics[width=1\textwidth]{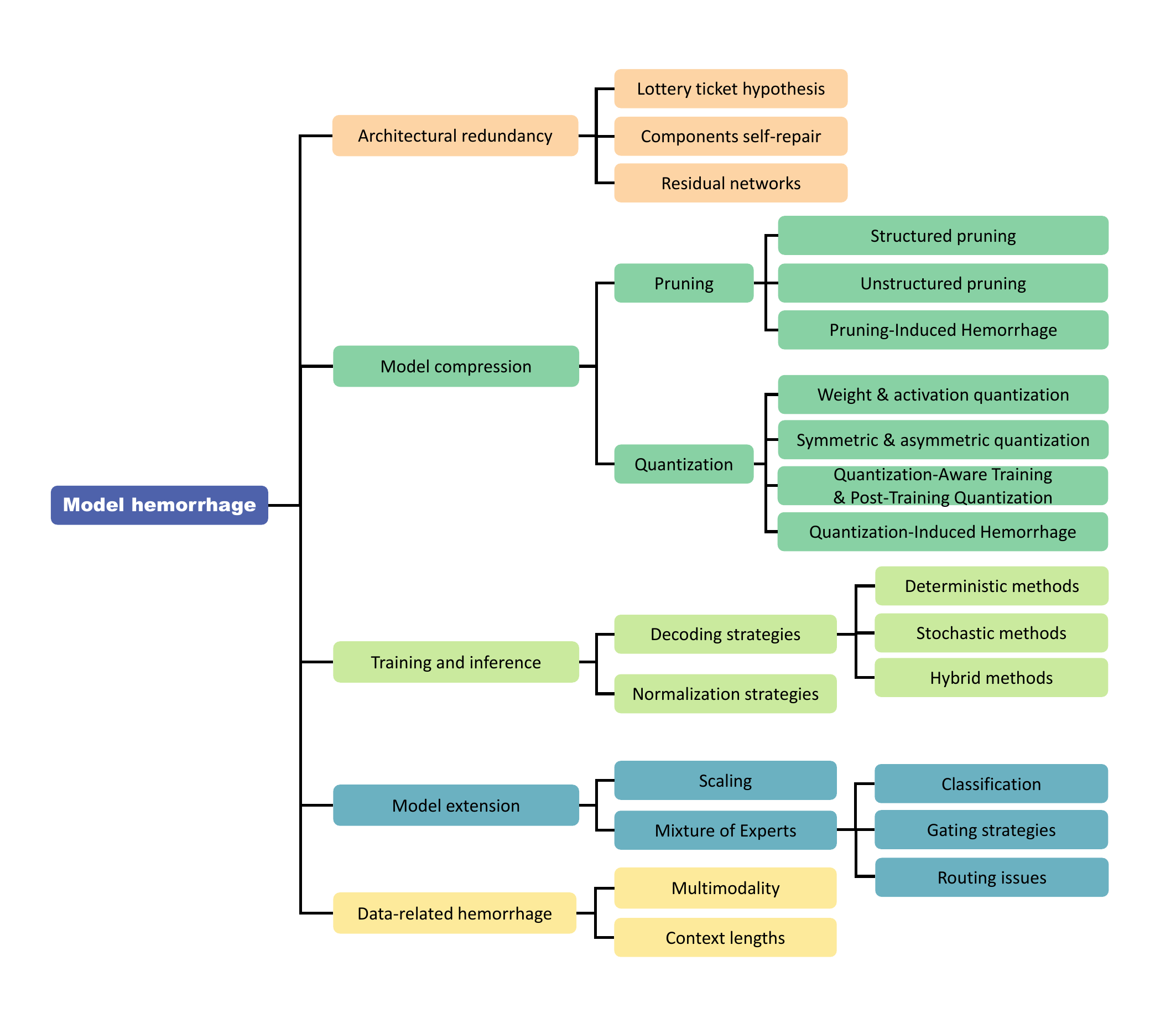} 
    \caption{The structural taxonomy for Model hemorrhage.}
    \label{fig:tree}
\end{figure*}
In this section, we will delve into the sources of robustness in large language models and the factors that contribute to the phenomenon of "Model Hemorrhage." We will explore the structural redundancy within the model, review operations such as pruning, quantization, decoding strategies, parameter expansion, and model framework selection, which may lead to model hemorrhage or performance degradation. Additionally, we will discuss the relationship between these operations and the model's robustness.
\subsection{Lottery Ticket Hypothesis and Architectural Redundancy}\label{sec:Redundancy}
The Lottery Ticket Hypothesis suggests that within the initial weight configurations of a large-scale neural network (the "lottery pool"), there exist certain subnetworks (the "winning tickets") that, through appropriate pruning operations, can achieve performance equal to or even surpassing the original network after retraining \cite{ref17}. By removing unnecessary components, these subnetworks retain the core structures crucial for robust performance. This hypothesis highlights that despite the redundancy present in large-scale networks, effective pruning can preserve the parts of the network essential for robustness. 

Furthermore, the hypothesis implies that the robustness of a neural network partly originates from those efficient substructures that inherently possess favorable characteristics \cite{ref18}. This has profound implications for understanding the role of network pruning in achieving both high performance and robustness, suggesting that robust networks may be built upon well-initialized subcomponents.

In some recent studies, The redundancy in large model architectures serves as the theoretical foundation for many optimization strategies \cite{SLEB, short_gpt, loraprune}, such as pruning, employed by researchers today. C. Rushing et al. \cite{self_repair} discovered a phenomenon of self-repair in large models, where downstream components compensate for the components pruned during pruning operations. When certain components, such as attention heads, MLP layers, or LayerNorm layers, are removed, the model exhibits a feedback-like repair mechanism, suggesting that there is structural redundancy within the model that maintains performance stability .

The redundancy in residual networks could be the core reason for the robustness of large models. In a residual network, the output is the weighted sum of the previous layer’s output and the current layer’s transformation. The mathematical formula is represented as:

\begin{equation}
x^{(\ell+1)} = x^{(\ell)} + f(x^{(\ell)}, \theta^{(\ell)})
\end{equation}
where \( x^{(\ell)} \) represents the output of the \( \ell \)-th layer, and \( f(x^{(\ell)}, \theta^{(\ell)}) \) is the transformation function of that layer. If the output of each layer is similar to that of the previous layers, then removing certain layer outputs will have a minimal impact on the final result. This explains the redundancy in deep networks: In deep networks, \( x^{(\ell)} \approx x^{(\ell-1)} + \epsilon \), meaning that pruning deep layers, except for the first and last layers, will not significantly affect the final output \cite{ref15}. Direct pruning of the final layer, however, can cause the model’s performance to collapse. A. Gromov et al. further demonstrated through cosine similarity-based pruning experiments that redundancy reduces the impact of removing certain layers on model performance. For example, some robust models, such as Llama-2-70B, maintain high MMLU accuracy even after 40\% of their layers are pruned \cite{unreasonable_ineffectiveness}.

\subsection{Pruning-Induced Model Hemorrhage}
\subsubsection{Structured Pruning}
Structured pruning involves removing neurons, attention heads, channels, sub-layers, or layers at different levels based on specific rules, or zeroing out weights in blocks proportionally (Semi-structured pruning). Structured pruning retains the overall network structure, making it more conducive to hardware acceleration. As noted in the work of X. Zhu et al. \cite{ref28}, structured pruning strategies can be categorized into three types based on pruning criteria and optimization objectives: Size-based Pruning, Regularization-based Pruning, and Loss-based Pruning.\\
\begin{figure*}[ht]
    \centering
    \includegraphics[width=1\textwidth]{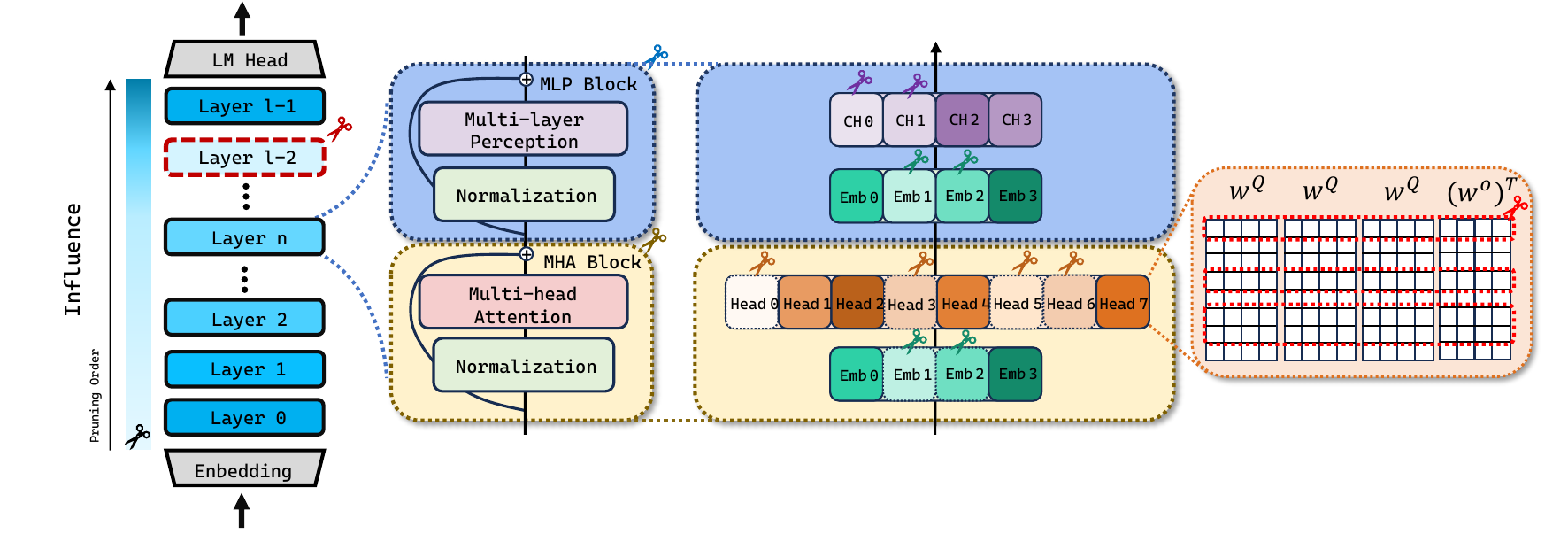}
    \caption{Structured pruning at different levels.}
    \label{fig:structured_pruning}
\end{figure*}\\
Size-based Pruning removes less important components by measuring the importance of weights, activations, or redundancy (e.g., weight norms or activation fluctuations) with the goal of directly reducing the model size while maintaining performance. Methods like FLAP \cite{flap_pruning} and ShortGPT \cite{short_gpt} fall under this category. Regularization-based Pruning introduces regularization terms (e.g., \(L_1\) regularization or angular distance regularization) into the objective function to constrain the weight distribution, inducing sparsity and selectively removing unimportant components. Examples include Sheared LLaMA \cite{sheared_llama} and SRAD \cite{unreasonable_ineffectiveness}. Loss-based Pruning quantifies the sensitivity of weights to the loss function to assess the impact of pruning on the overall model performance, prioritizing the removal of components that have minimal effects on the loss. This approach is exemplified by methods like LLM-Pruner \cite{llm_pruner} and SLEB \cite{SLEB}.

These three pruning strategies offer unique advantages and collectively support the goal of enhancing efficiency and robustness in large-scale models. Table~\ref{tab:pruning_methods} summarizes some structured pruning methods.

\definecolor{methodrow}{RGB}{216,214,194}
\definecolor{otherrow}{RGB}{236,234,223}

\begin{table}[h!]
\centering
\renewcommand{\arraystretch}{1.5}
\setlength{\aboverulesep}{0pt}
\setlength{\belowrulesep}{0pt}
\caption{Summary of Structured Pruning Methods, Formulas, and Categories.}
\begin{tabular}{ccc}
\toprule
\rowcolor{methodrow} \textbf{Method} & \textbf{Pruning Formula} & \textbf{Category} \\ 
\arrayrulecolor{gray}\specialrule{0.5pt}{0pt}{0pt}

\multirow{2}{*}{LLM-Pruner \cite{llm_pruner}}
  & \( W_{\text{pruned}} = W \cdot 1_{\text{Importance}_{|W|} \geq \tau} \) & \multirow{2}{*}{Loss-based} \\ 
  & \( \text{Importance}_{|W|} = \left| \frac{\partial L}{\partial W} W - \frac{1}{2} W^T H W \right| \) &  \\ 
\specialrule{0.5pt}{0pt}{0pt}

\rowcolor{otherrow}Shortened LLaMA \cite{shortened_llama} 
  & \( \text{Keep}(L) = \text{ArgMax}_{L} \frac{1}{|L|} \sum_{i \in L} \|W_i\| \) & Size-based \\ 
\specialrule{0.5pt}{0pt}{0pt}

\multirow{2}{*}{SLEB \cite{SLEB}} 
  & Prune block $j$ if $\text{Metric}_j^3(M')$ is minimized: & \multirow{2}{*}{Loss-based} \\
  & \( \text{Metric}_j^3(M') = - \frac{1}{K} \sum_{k=0}^{K} \log p_{M_j'}(w_k | w_{<k}) \) &  \\ 
\specialrule{0.5pt}{0pt}{0pt}

\rowcolor{otherrow}SliceGPT \cite{slice_gpt} 
  & \( W_{\text{slice-pruned}} = W_{\text{slice}} \cdot (1 - \beta) \) & Size-based \\ 
\specialrule{0.5pt}{0pt}{0pt}

\multirow{2}{*}{FLAP \cite{flap_pruning}}
  & \( W_{\text{pruned}} = W \cdot 1_{\Delta > \tau} \) & \multirow{2}{*}{Size-based} \\ 
  & \( \Delta = \text{Var}(\text{Activation}_{i,j}) \) &  \\ 
\specialrule{0.5pt}{0pt}{0pt}

\rowcolor{otherrow}Sheared LLaMA \cite{sheared_llama} 
  & \( W_{\text{pruned}} = W \cdot \left(1 - \alpha \cdot \frac{\|W\|}{\|W_{\text{max}}\|}\right) \) & Regularization-based \\ 
\specialrule{0.5pt}{0pt}{0pt}

\multirow{2}{*}{ShortGPT \cite{short_gpt}}
  & \( \text{BI}_i = 1 - \frac{X_i^T X_{i+1}}{\|X_i\|_2 \|X_{i+1}\|_2} \) & \multirow{2}{*}{Size-based} \\
  & \( W_{\text{shortened}} = W_{\text{original}} \cdot (1 - \rho) \) &  \\ 
\specialrule{0.5pt}{0pt}{0pt}

\rowcolor{otherrow}SRAD \cite{unreasonable_ineffectiveness} 
  & \( \Delta\theta = \frac{1}{\pi} \arccos\left(\frac{W_1 \cdot W_2}{\|W_1\| \|W_2\|}\right) \), Prune if \( \Delta\theta \leq \epsilon \) & Regularization-based \\ 
\bottomrule
\end{tabular}
\label{tab:pruning_methods}
\end{table}

\subsubsection{Unstructured Pruning}

Unstructured pruning is an optimization technique that achieves model sparsity by evaluating the importance of individual weights. Its flexibility and high compression rates make it a key method for optimizing large language models (LLMs). Unstructured pruning can achieve extremely high compression rates; for instance, Wanda achieves a 60\% sparsity rate on LLaMA-7B with minimal performance degradation across multiple downstream tasks~\cite{wanda}, while Flash-LLM achieves a 70\% sparsity rate on OPT-175B, significantly reducing storage requirements with less than 2\% performance degradation during inference~\cite{flashllm}. However, unstructured pruning often results in irregular sparse patterns in the weight matrix, necessitating specialized hardware accelerators (e.g., sparse matrix multiplication units) to efficiently handle sparse matrix computations and fully exploit the benefits of sparsity in terms of storage and computation.

Among various unstructured pruning methods, Magnitude Pruning is the most basic, directly removing weights with small magnitudes. While simple to implement, it does not account for the contextual importance of weights. SparseGPT~\cite{sparsegpt}, on the other hand, introduces a diagonal Hessian approximation to assess the impact of weights on errors, enabling more precise pruning at the cost of high computational complexity and hardware resource requirements.  Wanda~\cite{wanda} simplifies the SparseGPT algorithm by eliminating the need for Hessian approximations and instead computes pruning metrics by multiplying weights with input activations. This simplification significantly reduces computational complexity while achieving a balance between high accuracy and efficiency. Following this approach, many subsequent methods use SparseGPT and Wanda as baselines or build upon their foundations. RIA \cite{RIA} introduces a post-training pruning method that re-evaluates the importance of each weight element based on all input and output connections. ADMM \cite{admm} builds on SparseGPT by incorporating the Alternating Direction Method of Multipliers (ADMM) to restore model performance after pruning, using a simple iterative mask selection process for pruning. OWL \cite{yin2024outlier} integrates both Wanda and SparseGPT, proposing the OWL metric to allocate varying pruning rates across different layers. Similarly, BESA \cite{BESA} refines pruning by considering each transformer block's pruning error and allocating sparsity in a differentiable way, overcoming the perturbations associated with traditional layer-wise approaches. DsnoT \cite{dsnoprune} is also an extension of the SparseGPT and Wanda pruning strategies, introducing a training-free fine-tuning approach that iteratively refines sparse LLMs by adjusting sparse masks, minimizing the reconstruction error between sparse and dense models. Several pruning methods are developed independently of Wanda and SparseGPT. For example, Flash-LLM~\cite{flashllm} introduces a "Load-as-Sparse, Compute-as-Dense" strategy, which optimizes memory bandwidth while allowing tensor cores to perform computations as if the model were dense. LoRAPrune~\cite{loraprune} incorporates LoRA (Low-Rank Adaptation) modules to evaluate the importance of weights and activations, excelling in task-specific pruning scenarios, albeit at the expense of additional computational overhead due to the extra modules. Table~\ref{tab:un_pruning_methods} summarizes the specific details of these methods.

\begin{table}[h!]
\centering
\renewcommand{\arraystretch}{1.5}
\setlength{\aboverulesep}{0pt}
\setlength{\belowrulesep}{0pt}
\rowcolors{2}{otherrow}{white}  % 从第二行开始交替颜色
\caption{Comparison of pruning algorithms for unstructured pruning in LLMs. \( S_{ij} \) represents the pruning metric. References are provided for each method.}
\begin{tabular}{ccccc}
\arrayrulecolor{gray!50}
\toprule
\rowcolor{methodrow}  % 标题行颜色
\textbf{Method} & \textbf{Weight Update} & \textbf{Calibration} & \textbf{Pruning Metric} \( S_{ij} \) & \textbf{Complexity} \\
\specialrule{0.5pt}{0pt}{0pt}

Magnitude & No & No & \( |W_{ij}| \) & \( O(1) \) \\
\specialrule{0.5pt}{0pt}{0pt}

SparseGPT \cite{sparsegpt} & Yes & Yes & \( \frac{|W_{ij}|^2}{\text{diag}(X X^T + \lambda I)_{jj}} \) & \( O(d_{\text{hidden}}^3) \) \\ 
\specialrule{0.5pt}{0pt}{0pt}

LoRAPrune \cite{loraprune} & Yes & Yes & \( \left\| \frac{\partial \mathcal{L}}{\partial \mathbf{B}_{i,:}} \odot \mathbf{A}_{:,j} + \mathbf{B}_{i,:} \odot \frac{\partial \mathcal{L}}{\partial \mathbf{A}_{:,j}} \right\|_2^2 \cdot \left( W_{i,j} + (\mathbf{B}\mathbf{A})_{i,j} \right)^2 \) & \( O(d_{\text{hidden}}^2) \) \\
\specialrule{0.5pt}{0pt}{0pt}

Wanda \cite{wanda} & No & Yes & \( |W_{ij}| \cdot \|X_j\|_2 \) & \( O(d_{\text{hidden}}^2) \) \\
\specialrule{0.5pt}{0pt}{0pt}

DsnoT \cite{dsnoprune} & No & Yes & \(\mathbb{E}[W_{ij} \cdot X_j] \cdot \frac{1}{\text{Var}(X_j)}\) & \( O(d_{\text{hidden}}^2) \) \\
\specialrule{0.5pt}{0pt}{0pt}

Flash-LLM \cite{flashllm} & Yes & Yes & Load-as-Sparse, Compute-as-Dense & \( O(d_{\text{hidden}}^2) \) \\
\specialrule{0.5pt}{0pt}{0pt}

RIA \cite{RIA} & No & Yes & \( \left( \frac{|W_{ij}|}{\sum_k |W_{kj}|} + \frac{|W_{ij}|}{\sum_k |W_{ik}|} \right) \cdot \|X_j\|_2^{0.5} \) & \( O(d_{\text{hidden}}^2) \) \\
\specialrule{0.5pt}{0pt}{0pt}

ADMM \cite{admm} & Yes & Yes & \( |W_{ij}| \cdot \|X_j\|_2 \) & \( O(d_{\text{hidden}}^3) \) \\
\specialrule{0.5pt}{0pt}{0pt}

OWL \cite{yin2024outlier} & No & Yes & \( \text{Outlier Ratio} \propto |W_{ij}| \cdot \|X_j\|_2 \) & \( O(d_{\text{hidden}}^2) \) \\
\bottomrule
\end{tabular}
\label{tab:un_pruning_methods}
\end{table}

\subsubsection{Model Sensitivity to Pruning.}
Our systematic evaluation of pruning methods reveals critical insights into model robustness under structural compression. Structured pruning triggers immediate performance degradation, with model collapse occurring at approximately 50\% sparsity, even with post-pruning fine-tuning (see \ref{sec:structured_pruning_exp}). While such methods theoretically enable inference acceleration, practical gains diminish significantly beyond this threshold, yielding less than a 2× speedup at the cost of catastrophic accuracy loss. In contrast, unstructured pruning demonstrates superior robustness at moderate compression rates. Methods like SparseGPT and Wanda maintain stable performance below 60\% sparsity, with minimal perplexity degradation on WikiText2 (\ref{sec:unstructured_pruning_exp}). However, aggressive compression universally induces model collapse, mirroring the "cardinal sparsity" phase transition \cite{jaiswalcompressing,jaiswal2023emergence}. Notably, semi-structured pruning exhibits higher sensitivity than unstructured counterparts under equivalent sparsity, incurring greater accuracy loss for comparable computational benefits (\ref{sec:semi-structured_pruning_exp}). These findings emphasize the inherent trade-off between structural regularity and robustness, favoring unstructured approaches for performance-critical deployments.

\subsection{Quantization and Precision-Driven Hemorrhage}

Quantization aims to reduce the precision of model parameters, thereby decreasing storage and computational complexity, significantly improving inference efficiency and hardware compatibility. Specifically, quantization converts floating-point values (e.g., FP32) into fixed-point or integer values (e.g., INT8, FP16), effectively reducing the computational load and memory consumption during inference. Studies have shown that classical models such as AlexNet and ResNet, when quantized to INT8, can still achieve classification accuracy close to floating-point precision on the ImageNet dataset, demonstrating the effectiveness of quantization~\cite{jacob2018quantization}.

\subsubsection{Quantization Fundamentals}
\textbf{Weight Quantization and Activation Quantization.}
Weight Quantization and Activation Quantization are two fundamental directions in quantization. Weight quantization converts neural network weights from high-precision floating-point numbers to lower-precision integers (e.g., INT8 or INT4), reducing storage requirements and significantly lowering inference power consumption. Activation quantization further reduces memory usage and bandwidth requirements by quantizing intermediate activation values.

The distribution of weights and activations plays a critical role in determining quantization precision. For instance, many neural networks exhibit normally distributed or sparse weights, enabling effective performance retention even after clipping outliers or redistributing value ranges (e.g., through symmetric quantization).\\
\textbf{Symmetric and Asymmetric Quantization.}
In symmetric quantization, the quantization intervals for weights and activations are symmetric around zero, while asymmetric quantization allows non-symmetric intervals, which are more effective for complex data distributions. For example, the LSQ (Learned Step Size Quantization) method dynamically learns the quantization step size and adjusts strategies based on the actual distribution of weights and activations, thereby improving the adaptability of low-precision quantization~\cite{lsq2020}.
\begin{figure*}[ht]
    \centering
    \includegraphics[width=1\textwidth]{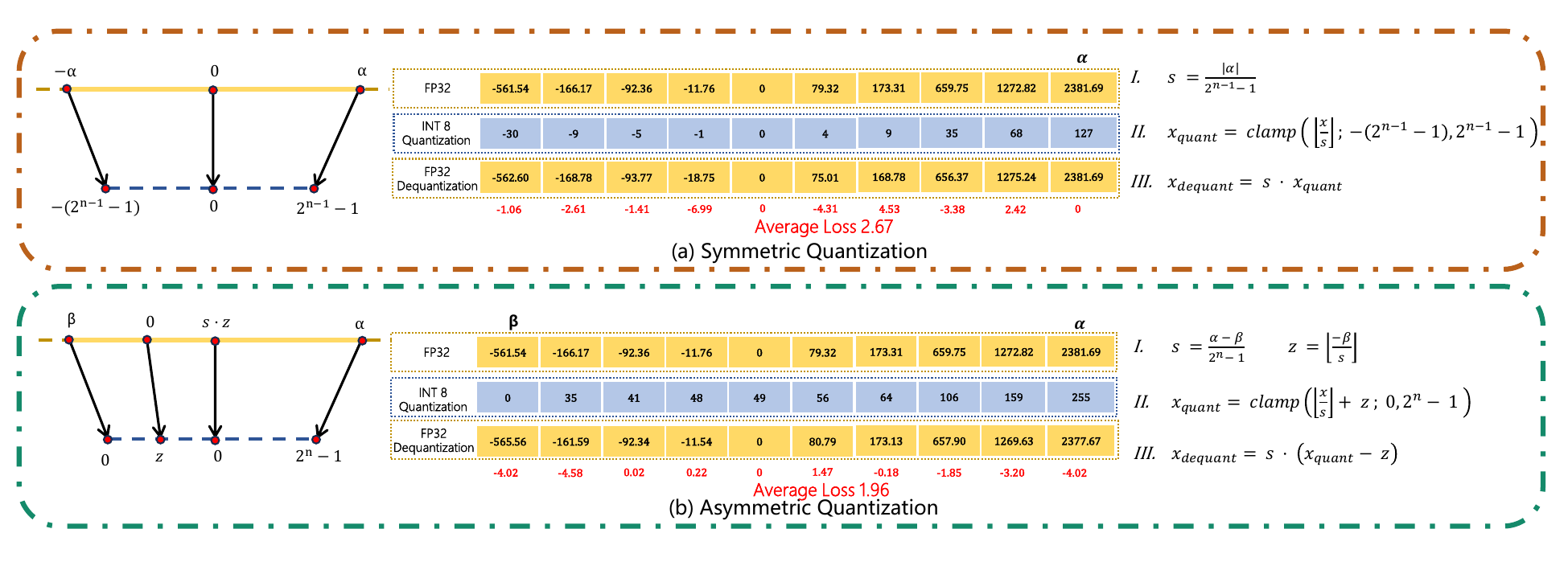} 
    \caption{Comparison of 8-bits Symmetric and Asymmetric Quantization.}
    \label{fig:sym_quant}
\end{figure*}
\subsubsection{Precision Restoration}
\textbf{Quantization-Aware Training (QAT).}
Quantization-Aware Training (QAT) is an optimization strategy that introduces simulated quantization noise during training to adapt models to quantization errors. Studies have shown that introducing quantization noise can act as a form of regularization, akin to data augmentation or Dropout, thereby enhancing model robustness \cite{liu2024llm}. For instance, simulating quantization errors during training significantly improves a model’s adaptability to low-precision computations in inference~\cite{quantization_survey2024}. Additionally, HAQ (Hardware-Aware Automated Quantization) uses reinforcement learning to automatically determine the optimal quantization bit-width for each layer, balancing resource utilization and performance~\cite{haq2019}.\\
\textbf{Representative Post-Training Quantization Techniques (PTQ).}
PTQ refers to the quantization process performed after the model has been trained, where the trained model is converted to a low-precision format (such as integers) to reduce memory usage and computational overhead, while attempting to maintain model performance. Representative PTQ techniques are shown in \ref{tab:quant_methods_v5}. For the post-training quantization needs of Transformer models, GPTQ \cite{gptq} was the first to achieve 3-4 bit low-precision compression while maintaining high performance. GPTQ uses a greedy strategy to quantize the Transformer model layer by layer, adjusting quantization thresholds using second-order gradient information to reduce error propagation. Its advantage lies in supporting high-precision, fast quantization of large-scale models on a single GPU (e.g., OPT-175B), achieving full quantization in just 4.2 hours, with a PPL performance loss of only 1-3\% after 4-bit quantization, enabling the deployment of extremely large models on a single A800 GPU. However, the GPTQ quantization process depends on GPU computation, requiring high hardware specifications, and the quantized model format has limited compatibility with some inference frameworks. AWQ \cite{awq} is an adaptive quantization method that emphasizes joint optimization of weights and activations, identifying critical weights through activation statistics and dynamically adjusting the quantization granularity. AWQ achieves lower accuracy loss than GPTQ under the same bit-width, but its quantization process has higher computational costs and requires additional calibration data. Traditional quantization operations rely on GPU hardware. To achieve efficient quantization and inference on CPU hardware platforms, GGML utilizes AVX/NEON instruction sets to accelerate low-precision computations, making large model quantization feasible in low-resource environments. However, due to poor file format scalability and lack of metadata support, it was later replaced by GGUF. As an upgraded version of GGML, GGUF solves the fragmentation problem of formats, integrates AVX instruction set acceleration and CUDA computation, and supports multi-hardware extensions. GGUF supports various quantization precisions from 1-bit to 8-bit, and under extreme compression, it can reduce the model file size of the 671B ultra-large DeepSeek-R1 \cite{deepseek_r1} model to below 140GB.
\begin{figure*}[ht]
    \centering
    \includegraphics[width=1\textwidth]{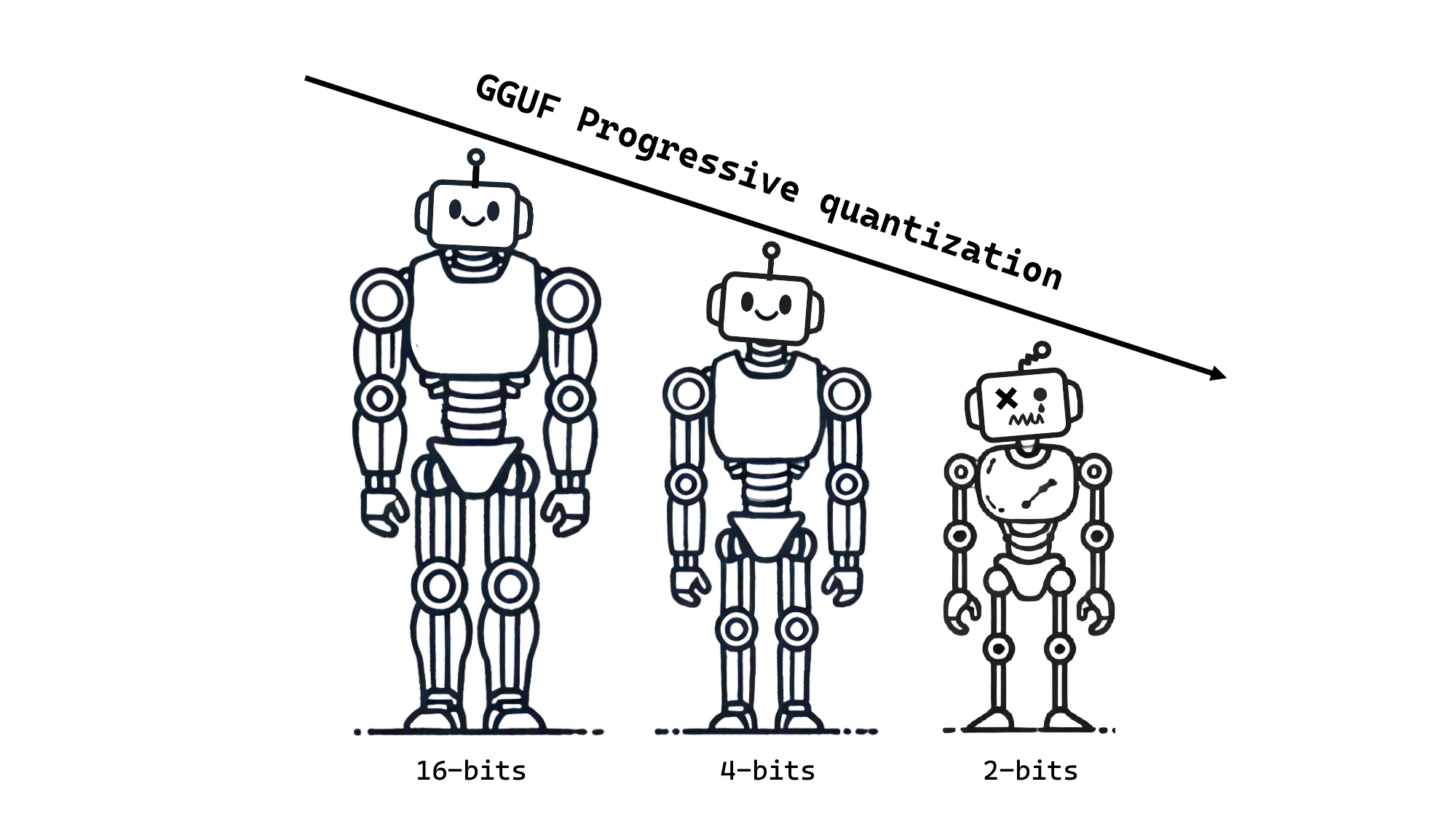} 
    \caption{GGUF Progressive quantization.}
    \label{fig:sym_quant}
\end{figure*}
% \begin{table*}[ht]
%   \centering
%   \caption{Evolution of LLM Quantization Methods with Hardware Support Breakdown}
%   \label{tab:quant_methods_v5}
%   \footnotesize
%   \begin{adjustbox}{width=1\textwidth} % 调整表格宽度以减少空白
%   \begin{tabularx}{\linewidth}{@{}l X X c c c@{}}
%     \toprule
%     \textbf{Method} & 
%     \textbf{Motivation/Background} & 
%     \textbf{Key Technique} & 
%     \textbf{Bit Width} & 
%     \multicolumn{2}{c}{\textbf{Hardware Support}} \\
%     \cmidrule(lr){5-6}  % 子列分隔线
%     & & & & 
%     \textbf{Quantization} & 
%     \textbf{Inference} \\
%     \midrule
    
%     GPTQ \cite{gptq} & 
%     Post-training quantization without fine-tuning & 
%     Layer-wise Hessian optimization; Adaptive rounding; Single-GPU workflow & 
%     2-4, 8 & 
%     GPU & 
%     CPU/GPU \\
%     \addlinespace[0.3em]
    
%     AWQ \cite{awq} & 
%     Joint optimization of weights and activations & 
%     Activation-value sensitivity; RL-based bit allocation; Mixed-precision protection & 
%     3, 4, 8 & 
%     GPU & 
%     GPU \\
%     \addlinespace[0.3em]
    
%     GGML & 
%     CPU-based edge device inference & 
%     Fixed-block quantization; AVX/NEON optimization; Memory-mapped loading & 
%     4-8 & 
%     CPU & 
%     CPU \\
%     \addlinespace[0.3em]
    
%     GGUF & 
%     Unified format for multimodal support & 
%     Dynamic block sizing; Hybrid tensor precision; CUDA/OpenCL backends & 
%     1-8 & 
%     CPU/GPU & 
%     CPU/GPU \\
%     \bottomrule
%   \end{tabularx}
%   \end{adjustbox}
% \end{table*}
\begin{table*}[ht]
  \centering
  \caption{Evolution of LLM Quantization Methods}
  \label{tab:quant_methods_v5}
  \footnotesize
  \setlength{\aboverulesep}{0pt}
  \setlength{\belowrulesep}{0pt}
  \begin{adjustbox}{width=1\textwidth}
  \begin{tabularx}{\linewidth}{@{}l X X c@{}}
    \toprule
    \rowcolor{methodrow}
    \textbf{Method} & 
    \textbf{Motivation/Background} & 
    \textbf{Key Technique} & 
    \textbf{Bit Width} \\
    \arrayrulecolor{gray!50}\specialrule{0.4pt}{0pt}{0pt}
    
    GPTQ \cite{gptq} & 
    Post-training quantization without fine-tuning & 
    Layer-wise Hessian optimization; Adaptive rounding; Single-GPU workflow & 
    2-4, 8 \\
    \noalign{\vspace{-1pt}}
    
    \rowcolor{otherrow}
    AWQ \cite{awq} & 
    Joint optimization of weights and activations & 
    Activation-value sensitivity; RL-based bit allocation; Mixed-precision protection & 
    3, 4, 8 \\
    \noalign{\vspace{-1pt}}
    
    GGML & 
    CPU-based edge device inference & 
    Fixed-block quantization; AVX/NEON optimization; Memory-mapped loading & 
    4-8 \\
    \noalign{\vspace{-1pt}}
    
    \rowcolor{otherrow}
    GGUF & 
    Unified format for multimodal support & 
    Dynamic block sizing; Hybrid tensor precision; CUDA/OpenCL backends & 
    1-8 \\
    \bottomrule
  \end{tabularx}
  \end{adjustbox}
\end{table*}

\subsubsection{Model Sensitivity to Quantization.}
Different network architectures exhibit varying sensitivities to quantization. Certain layers, such as initial and output layers, are highly sensitive to quantization errors and require higher precision representation, whereas some intermediate layers, such as the Feedforward layers in Transformers, are more robust to low-precision quantization. For example, studies have shown that Attention layers in Transformers are particularly sensitive to precision degradation during quantization, while Feedforward layers maintain acceptable performance even after INT8 quantization. Similarly, in vision models, shallow convolutional kernels are more sensitive to quantization errors than deeper feature mappings~\cite{zeroquant2022}.Further investigations from ZeroQuant-V2 reveal that activation quantization is generally more sensitive than weight quantization, especially in larger models. For models exceeding 10B parameters (e.g., OPT-66B \cite{zhang2022opt} and BLOOM-176B \cite{bloom}), activation quantization leads to more significant accuracy degradation compared to smaller models~\cite{zeroquant_v2}. This phenomenon highlights the need for careful handling of activations during quantization in large-scale language models.

Our experiments reveal systematic patterns in model robustness under progressive precision reduction. A critical 3-bit threshold emerges, beyond which accuracy degradation transitions from linear to exponential collapse—evident in both perplexity (WikiText-2) and reasoning benchmarks (ARC-C, MMLU). Low-bit quantization (2-bit) achieves 5× parameter compression while still retaining >90\% of baseline performance  (Table \ref{tab:full-results}). Quantization universally outperforms pruning in robustness, sustaining <5\% accuracy loss at 4-bit compression versus unstructured pruning’s 15-20\% degradation under matched sparsity (\ref{sec:quantization_comp}). This advantage amplifies at extreme compression (70\%+), where quantization preserves functional coherence while pruning induces hemorrhage.
%\textbf{Impact of Quantization Noise.}
%Quantization noise is theoretically treated as a form of random perturbation during inference, similar to the effects of data augmentation. It enhances model robustness to small perturbations while allowing the model to adjust automatically to quantized inference requirements through QAT. Studies show that QAT-trained models exhibit superior stability in low-precision inference scenarios. For example, QAT has demonstrated significant effectiveness in improving time-series prediction for Recurrent Neural Networks~\cite{quantization_survey2024}.

\subsection{Decoding Strategies and Performance Degradation}

Decoding methods are the critical bridge connecting the probability distributions predicted by large language models (LLMs) for the next token to the actual generated sequences. Selecting an appropriate decoding strategy significantly impacts model performance, generation quality, and robustness, especially in closed-ended tasks (e.g., machine translation, question answering) and open-ended tasks (e.g., dialogue generation, creative writing). Decoding methods are broadly categorized into deterministic and stochastic approaches.

\subsubsection{Deterministic Methods}

Deterministic methods generate text by searching for globally optimal or suboptimal solutions within the output probability distribution, prioritizing high accuracy and logical consistency. These methods excel in closed-ended tasks that require precision, such as code generation and question answering. Basic approaches include Greedy Search, which selects the highest-probability token at each step but risks repetitive outputs, and Beam Search (BS), which retains multiple candidate paths to optimize cumulative probability, significantly improving coherence \cite{holtzman2020curious, freitag2017beam}.

Advanced deterministic methods extend beam search to balance quality and diversity. For instance, Diverse Beam Search (DBS) introduces diversity penalties to avoid redundant outputs \cite{vijayakumar2018diverse}, while Contrastive Search (CS) optimizes contrastive scores to suppress generic or inconsistent tokens \cite{su2022contrastive}. These extensions enhance performance in content-generation tasks like summarization and narrative generation, where precision and semantic diversity are critical.

\subsubsection{Stochastic Methods}

Stochastic methods enhance diversity and creativity in generated content by introducing randomness into the decoding process. These methods are better suited for open-ended tasks, such as dialogue generation and creative writing. Sampling strategies include Top-p Sampling, which dynamically selects tokens based on cumulative probability thresholds to avoid verbosity \cite{holtzman2020curious}; Top-k Sampling, which restricts sampling to the top \( k \) tokens for quality-diversity balance \cite{hierarchical_story_generation}; and Temperature Sampling, where adjusting the temperature parameter \( \tau \) modulates distribution sharpness, trading coherence for diversity at higher \( \tau \) values.

Advanced stochastic strategies optimize sampling through dynamic adaptation. For example, Mirostat \cite{basu2021mirostat} adjusts sampling ranges to maintain target perplexity levels, while Typical Sampling \cite{meister2023typical} prioritizes tokens based on entropy to align with human language patterns. These methods systematically address the trade-off between fluency and unpredictability, ensuring outputs remain both engaging and contextually plausible.
\subsubsection{Hybrid Methods}

Hybrid methods integrate deterministic search with stochastic sampling to dynamically balance controllability and diversity, addressing limitations of purely deterministic (repetitive patterns) and purely stochastic (incoherence) approaches. By combining global optimization capabilities with localized randomness, they achieve flexible control over generation quality, making them suitable for scenarios requiring both consistency and creativity, such as technical writing or interactive storytelling. A representative example is Stochastic Beam Search \cite{gumbel_topk}, which injects Gumbel noise into beam search to diversify candidates while retaining global optimization through probabilistic path selection.

Speculative decoding implementations based on speculative sampling exemplify this hybrid paradigm. During the drafting phase, a lightweight draft model generates candidate sequences via stochastic sampling strategies, introducing localized randomness to explore diverse hypotheses. The verification phase then employs the target model to perform parallel scoring and deterministic probability calibration across all candidates. This hierarchical process ensures that the final outputs remain statistically unbiased and distributionally aligned with the target model while leveraging the draft model's efficiency to reduce sequential inference steps \cite{lookahead_decoding, leviathan2023fast, chen2023accelerating, li2024eagle}.

\subsubsection{Performance, Speed, and Robustness}
\textbf{Performance}: Deterministic methods generally excel in closed-ended tasks, where logical consistency and accuracy are critical. For instance, beam search consistently outperforms stochastic methods in datasets like HumanEval \cite{chen2021evaluating} and GSM8K. Conversely, stochastic methods, such as Top-p and temperature sampling, excel in open-ended tasks by generating diverse and creative content but may sacrifice semantic coherence \cite{holtzman2020curious}.\\\\
\textbf{Speed}: Stochastic methods and efficient deterministic methods (e.g., FSD, Greedy Search) provide substantial advantages in real-time applications due to their low computational complexity. Contrastive decoding introduces a 1.4x latency compared to greedy search due to its reliance on an auxiliary model but maintains consistent latency across varying sequence lengths \cite{decoding_methods_llms}, making it suitable for long-sequence generation tasks \cite{su2022contrastive}. In contrast, beam search and diverse beam search exhibit linearly increasing latency with sequence length, illustrating a trade-off between speed and quality. Speculative decoding fundamentally redefines this balance through hybrid architectures: By integrating stochastic draft generation (via smaller models or intermediate feature prediction) with parallelized target model verification, it achieves 2-3× acceleration over autoregressive baselines while preserving distributional fidelity \cite{leviathan2023fast, lookahead_decoding, li2024eagle, cai2024medusa}. Advanced implementations like EAGLE-3 \cite{eagle3} further optimize draft-target alignment through multi-layer feature fusion and training-time test techniques, claiming up to 6× speedup ratios on code generation and mathematical reasoning tasks.\\\\
\textbf{Sensitivity to Decoding Methods.} Alignment optimization significantly enhances model robustness by reducing dependency on decoding methods. Aligned models, such as LLaMA2-7B-Chat \cite{touvron2023llama}, exhibit stable performance across various tasks and decoding strategies. Unaligned models, like LLaMA2-7B, display significant performance variation depending on the decoding method used, emphasizing the importance of alignment optimization for robust performance. Our systematic evaluation reveals significant task- and model-dependent variations in decoding strategy efficacy. Deterministic methods (e.g., FSD/FSD-d \cite{yang2023fsd}) exhibit robust performance on closed-ended tasks (MBPP, GSM8K), achieving <5\% accuracy degradation compared to stochastic baselines, yet underperform in open-ended generation due to constrained diversity (Table \ref{tab:performance}). Stochastic approaches demonstrate inverse trends—temperature sampling maintains stable quality across tasks, while $\eta$-sampling suffers pronounced instability ($\Delta$ >15\% PPL on Wikinews). For speculative decoding methods, Blockwise Parallel Decoding \cite{stern2018blockwise} introduces a block verification mechanism to break autoregressive dependencies. It achieves 2x speedup with no quality loss, or up to 7x acceleration with controlled quality degradation. DistillSpec \cite{zhoudistillspec} further optimizes draft-target alignment via knowledge distillation, improving speed by 10–45\% under greedy decoding while preserving output distribution. When allowing lenient verification, it reduces latency by 6–10x with negligible performance loss.

%\textbf{Hyperparameter Sensitivity}: The effectiveness of decoding methods depends heavily on hyperparameter settings, which vary across tasks and models. Deterministic methods, such as beam search, require careful tuning of beam width \( k \), while stochastic methods depend on parameters such as \( p \) and \( k \). Research indicates that methods like temperature sampling and contrastive decoding achieve superior task-specific performance when hyperparameters are well-tuned but experience significant drops under fixed settings. For instance, temperature sampling exhibits an 11.59\% performance decrease on LLaMA2-7B with fixed hyperparameters across datasets, highlighting its sensitivity to parameter selection \cite{decoding_methods_llms}. In contrast, methods such as FSD and FSD-d \cite{yang2023fsd} demonstrate low sensitivity to hyperparameters, delivering consistent performance across diverse tasks without extensive tuning.

\subsection{Normalization-Induced Hemorrhage in LLM Training}

The choice of normalization methods significantly impacts the performance and robustness of large language models (LLMs). In particular, layer normalization (LayerNorm) \cite{LayerNorm} has a pronounced effect on both model performance and training stability. Notably, LayerNorm is sometimes replaced with more computationally efficient RMSNorm \cite{RMSNorm} in practical training of large models to reduce computational overhead.

\begin{figure}[h]
    \centering
    \includegraphics[width=1\linewidth]{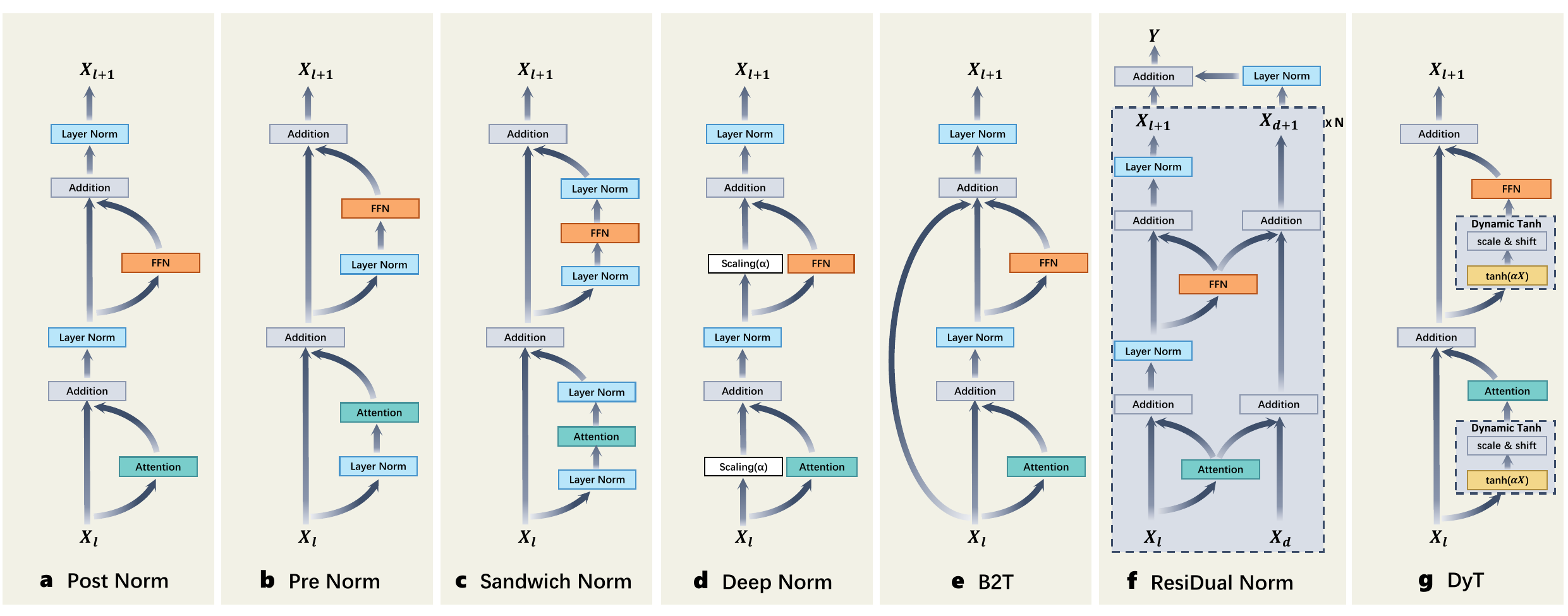}
    \caption{Illustration of Different LayerNorm Structures in Transformers.}
    \label{fig:Norm}
\end{figure}

\subsubsection{Gradient Vanishing vs. Representation Collapse}
LayerNorm, a widely used normalization method, has two main variants: Post-LN and Pre-LN. Post-LN applies normalization after each sub-layer, following the residual connection, and provides stronger regularization, often leading to better performance. However, it can cause gradient vanishing or exploding issues in very deep models due to its application across all parameters \cite{layernorm_transformer}. This makes Post-LN more suitable for relatively shallow models, such as BERT \cite{ref8}. In contrast, Pre-LN normalizes before each sub-layer, stabilizing training by avoiding gradient issues. This design, while providing greater stability, is more common in deeper models \cite{ref5, llama, baichuan2, deepseek_v3, qwen2.5, bloom}.

However, Pre-LN introduces a new challenge known as "representation collapse." By placing layer normalization within the residual block, it weakens the residual connections at deeper layers, causing representations to converge and reducing the ability to distinguish hierarchical features \cite{liu2020understanding}. Researchers have explored various solutions to address this limitation by altering weight initialization, adjusting training parameters during early stages, or redesigning residual structures. For example, T-fixup \cite{T-fixup} employs depth-dependent scaling of encoder and decoder weight matrices during initialization, decreasing dependence on warm-up phases. BranchNorm \cite{liu2023branchnorm} stabilizes gradients in early training by dynamically adjusting scaling factors in non-residual branches. Similarly, B2T \cite{B2T} introduces additional residual connections after every two sub-layers based on the Post-LN structure, facilitating direct gradient propagation through residual paths and preventing gradient vanishing. ResiDual \cite{xie2023residual} integrates advantages of both Pre-LN and Post-LN structures, constructing a hybrid Pre-Post-LN architecture that capitalizes on their respective strengths. Additionally, CogView \cite{sandwich} combines characteristics of both variants by proposing Sandwich-LN, a method that adds LayerNorm before and after the feed-forward network (FFN) and attention heads within residual branches to regularize intermediate representations effectively, as seen in Gemma-2 \cite{Gemma2024}. Similar to Sandwich-LN, Sub-LN \cite{wang2022foundation} introduces two layers of normalization within each sub-module (Attention and FFN), which retains the stability of Pre-LN while enhancing the independence of sub-modules, effectively preventing representation collapse. Recent studies have even started exploring alternatives to the LayerNorm structure. Since the core function of LayerNorm is to scale inputs and compress extreme values, DyT \cite{Zhu2025DyT} controls the input scaling using a learnable parameter \( \alpha \) and employs the S-shaped curve of the tanh function to restrict the range of activation values. This allows stable training without introducing normalization.

\subsubsection{Challenges in Scaling Beyond 100B Parameters}
As model depths increase, researchers have observed that even Pre-LN structures encounter significant challenges when models exceed approximately 100 billion parameters. Large models often rely on mixed-precision training, such as FP16, to improve computational efficiency during forward and backward passes. However, Pre-LN architectures can lead to overflows (NaN) or loss diffusion (underflows), which hinder model stability. To address these issues, Sandwich Norm introduces an extra LayerNorm at the end of each residual branch and incorporates PB-Relax techniques to stabilize training by mitigating extreme attention values and removing bias terms.

Building on these observations, the GLM-130B team \cite{glm130b} identified instability in training large-scale models such as OPT-175B \cite{zhang2022opt} and BLOOM-176B \cite{bloom}. These models often experienced loss spikes that led to frequent model collapse and degraded performance. While OPT-175B mitigated these issues by manually skipping abnormal data and adjusting hyperparameters, this workaround did not fundamentally resolve the underlying causes of training instability. In contrast, BLOOM-176B added an extra LayerNorm after the embedding layer, effectively suppressing abnormal gradients from the embedding layer and improving stability. However, the GLM team found that this modification compromised the model’s zero-shot generalization capability. As a result, they concluded that Pre-LN provided no substantial advantages over Post-LN for models beyond the 100B parameter scale.

To address these challenges, the GLM-130B team adopted an enhanced Post-LN approach called DeepNorm. DeepNorm \cite{deepnet} improves upon traditional Post-LN by introducing scaling factors (\( \alpha \)) and initialization gains (\( \beta \)) into the residual connections, which helps limit signal growth and ensures more stable gradient flows in deep models. This design enables the training of transformers with up to 1,000 layers, making it a robust choice for extremely deep architectures while maintaining gradient stability. As a result, DeepNorm demonstrated superior performance on zero-shot tasks like LAMBADA compared to GPT-3-175B, OPT-175B, and BLOOM-176B.

\subsection{Scaling Paradox: From Growth to Hemorrhage}

The astonishing feature of large-scale models lies in their apparent ability to improve given sufficient data and computational resources. Intuitively, larger models should exhibit stronger capabilities. However, several studies indicate that blindly increasing the number of parameters or layers may degrade model robustness \cite{performance_law_llms}. A large model consuming more resources, equipped with more parameters yet delivering no significant performance improvement—or even experiencing performance degradation—resembles an individual suffering from obesity. This consideration underpins the inclusion of parameter scaling as a contributing factor in the "Model Hemorrhage" framework.

The history of model parameter scaling reveals that performance depends primarily on the number of parameters, but it is also influenced by dataset size and computational resources used during training. The mathematical relationship among these factors was elucidated by Kaplan et al \cite{kaplan2020scaling}. in their pioneering work, which spurred a wave of attention to "Scaling Laws." Numerous subsequent large-scale models have adhered to these training paradigms, including Gopher-280B, whose training methodology was directly inspired by Kaplan's findings. However, the paradigm underpinning Gopher's training is not without flaws. 

Does an increase in parameter count always benefit performance? The Gopher team discovered, through toxicity and bias analysis, that as the toxicity of a prompt increases, larger models are more likely to generate responses with higher toxicity. In essence, larger models amplify the adverse effects of toxic inputs. Recent studies corroborate this phenomenon \cite{poison_bench}, showing that certain models, such as Qwen-2.5, exhibit heightened sensitivity to injection attacks as their parameter count grows.

Further refinements to the scaling law were introduced by the Chinchilla-70B team, who argued that when the number of FLOPs used for training is fixed, model size and the amount of training data should scale proportionally. They introduced the concept of "Compute-Optimal LLMs," proposing a training strategy that prioritizes increasing the size of training datasets. Empirical results validated this approach: Chinchilla-70B, trained with a larger dataset, outperformed Gopher-280B on most tasks, despite having only one-quarter of Gopher's parameters.

How can we assess whether a model exhibits "Model Hemorrhage" during parameter scaling? The "Performance Law" offers a novel perspective \cite{performance_law_llms}. By comparing the predicted and actual MMLU performance of LLMs across varying parameter sizes and architectures, researchers can gauge the model's health status. If the predicted values significantly exceed the actual values, potential issues may arise from inadequate dataset size, suboptimal training strategies, or low computational precision. Conversely, if predicted values are markedly lower than actual values, it could indicate data leakage during training.

\subsection{Routing Vulnerabilities in Mixture-of-Experts}

\subsubsection{Classification of MoE Models}
\begin{figure*}[ht]
    \centering
    \includegraphics[width=0.95\textwidth]{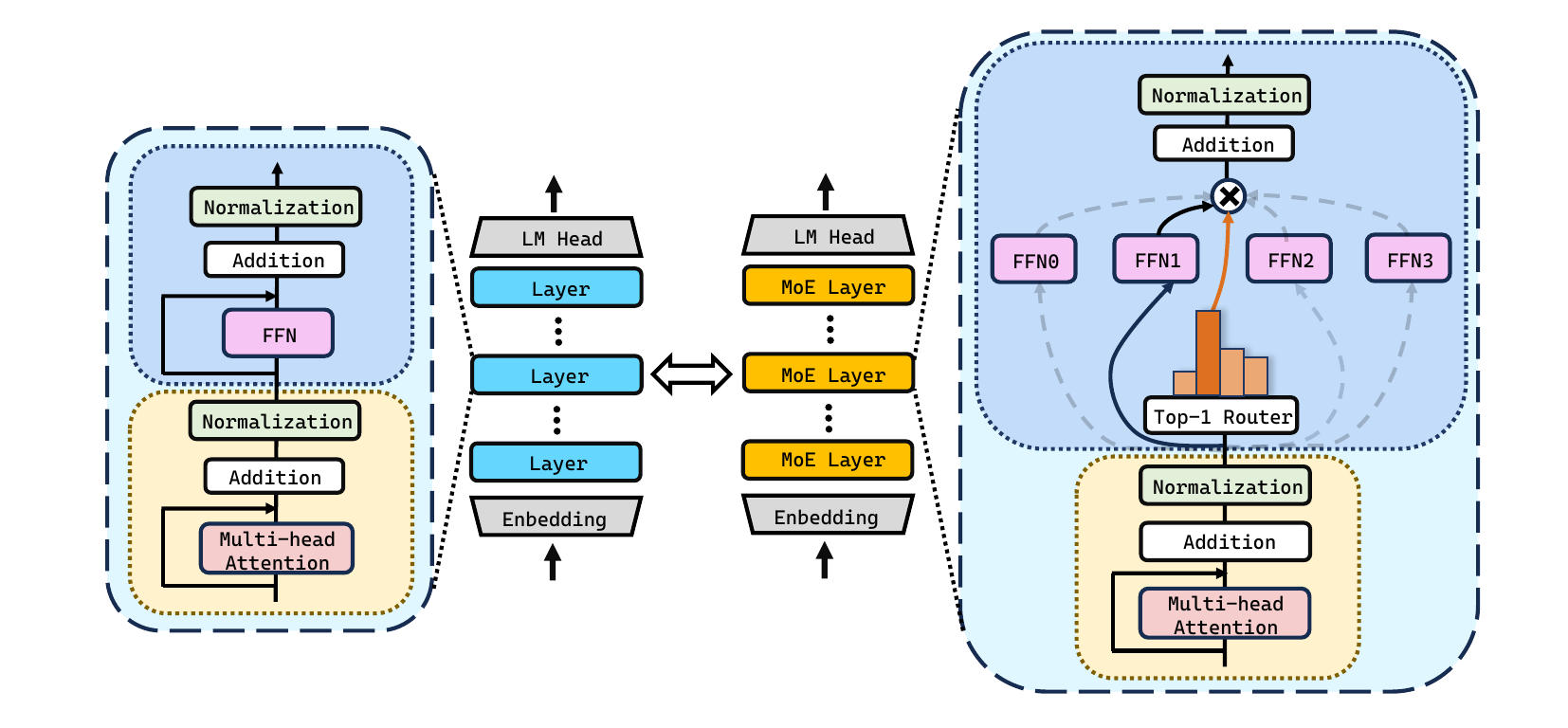}
    \caption{The comparison between the standard model and the MoE extended model with the Top-1 routing strategy.}
    \label{fig:MoE_expansion}
\end{figure*}

MoE serves as an extension to large language models (LLMs). In the Transformer architecture, each layer typically includes a fully connected feedforward network (FFN) \cite{cai2024survey}. MoE extends this by introducing multiple “expert” networks. Instead of passing inputs through a fixed FFN, the model routes the inputs to a subset of expert networks. Each expert can be seen as a sub-network designed to process specific tasks or data patterns.  The structural transformation of a standard Transformer layer into a MoE-augmented layer with Top-1 routing is illustrated in Figure~\ref{fig:MoE_expansion}.

MoE models can be broadly categorized based on how expert networks are activated during computation. Dense MoE involves all experts participating in computation, with the output being a weighted sum of all experts. This approach incurs significant computational costs, as seen in LoRAMoE~\cite{LoRAMoE}. Sparse MoE activates only a subset of experts to reduce computational overhead, making it the mainstream architecture in MoE models. Examples such as Switch Transformer~\cite{fedus2022switch} and GLaM~\cite{du2022glam} achieve efficient inference through sparse activation. However, they heavily rely on the robustness of the gating mechanism and are susceptible to input noise. The fully differentiable soft MoE adopts soft selection when choosing experts. Unlike traditional sparse MoE that activates only one or a few experts, this approach combines outputs from multiple expert networks using weighted summation. The weights for each expert are calculated using a continuous differentiable function, which can be optimized directly through gradient descent~\cite{SoftMoE}.

\subsubsection{Gating Strategies in Sparse MoE Architectures}
The Mixture of Experts (MoE) architecture employs a gating mechanism to dynamically activate experts based on input features, producing outputs as weighted sums of selected experts' contributions. This balances model scalability with computational efficiency, where robustness hinges on architectural design and gating strategies.

The Top-k gating strategy~\cite{Shazeer2017} activates the top-ranked k experts per input, enabling collaborative integration of multiple experts. For instance, GShard~\cite{GShard} uses top-2 gating, reducing single-expert dependency.Noise injection \( R_{\text{noise}} \) during gating training enhances exploration and stability~\cite{Shazeer2017,fedus2022switch}. Alternative approaches like Top-1 gating in M6-t~\cite{M6t} partition experts into groups with individual Top-1 selections, achieving comparable performance while improving efficiency.

To address challenges related to sparse expert allocation, Token Linear Balancing~\cite{BASELayers} formulates the token-to-expert assignment as a linear assignment problem, ensuring even distribution of tokens across experts and balancing computational load. Another gating strategy, The Random Routed Experts (RRE) mechanism~\cite{PanGu2} combines domain mapping and random gating strategies. Tokens are first routed to expert groups corresponding to specific domains and then randomly assigned within the group. Random routing enhances model diversity and mitigates over-reliance on specific experts. A further variation, Top-k Token Gating \cite{ExpertChoice} further refines this approach by enabling experts to select the top-k tokens they will process, ensuring balanced load distribution. Another innovative approach is the Attention Router, introduced in Yuan 2.0-M32\cite{YuanAttentionRouter}, which uses an attention mechanism to dynamically route tokens. In this approach, each expert computes a weighted sum of all input tokens based on attention scores. Finally, Soft MoE with Expert Merging \cite{SoftMoE} introduces an “expert merging” strategy, where outputs from multiple experts are combined using soft weights, offering a more flexible integration of expert contributions.

\subsubsection{Routing Issues in Mixture-of-Experts}

\textbf{Routing Fluctuation} refers to significant variations in token-to-expert assignments during training, which can lead to some experts being over-activated while others remain underutilized. This fluctuation not only reduces computational efficiency but also destabilizes model training~\cite{RectifyRouterMoE}. StableMoE~\cite{StableMoE} proposes a stable routing strategy using two-stage training. In the first stage, a balanced and coherent routing strategy is learned and distilled into a lightweight router. In the second stage, the router is frozen to ensure stability. Research such as GShard~\cite{GShard} and Switch Transformer~\cite{fedus2022switch} has introduced auxiliary load balancing losses to mitigate the negative impact of routing fluctuation.\\
\textbf{Expert Load Imbalance} occurs when certain experts process significantly more tokens than others, leading to uneven computational load distribution. This issue is commonly observed in greedy top-k routing strategies, resulting in both over-utilized and under-utilized experts. Overloaded experts are more prone to overfitting, while under-utilized experts waste computational resources. 

Load imbalance also manifests as dropped tokens and padding tokens. Due to expert capacity limits, some tokens may be dropped during routing, reducing the model's ability to learn from specific input features. Conversely, padding tokens are introduced when the number of tokens assigned to an expert is less than its capacity, resulting in redundant computations and reduced efficiency.

Rectify-Router~\cite{RectifyRouterMoE} addresses these issues through Intra-GPU Rectification, which reassigns dropped tokens, and Fill-in Rectification, which replaces padding tokens with high-scoring tokens, ensuring better load balancing.

\subsection{Data-related hemorrhage}

\subsubsection{Multimodality}
Recent studies have also explored the robustness of Vision Transformers (ViT) in various contexts, such as robustness against common corruptions \cite{Bhojanapalli2021}, distribution shifts \cite{Paul2022}, and adversarial attacks \cite{Mahmood2021, Mao2021, Aldahdooh2021}, with several benchmarks being proposed for evaluating robustness in these models \cite{Wenzel2022, zhang2024benchmarking}. Recent studies have begun to focus on the linguistic properties of prompts and their interference with model performance~\cite{leidinger2023prompting}. In terms of modality, declarative sentences have a more significant impact on model performance compared to interrogative and imperative sentences. Regarding synonyms, the use of non-standard synonyms unexpectedly improves model accuracy. Furthermore, some research~\cite{pezeshkpour2024sensitivity} has identified a "position bias" in large models, where the order of options in multiple-choice questions affects accuracy. Data leakage is also a notable source of interference; for example, the numerical reasoning ability of models is significantly influenced by the frequency of relevant numerical terms appearing in training datasets~\cite{razeghi2022numerical}. Adversarial GLUE \cite{Wang2021} focuses on adversarial attacks targeting large language models.

\subsubsection{Context lengths}
A widely studied phenomenon in long-context processing is the “Lost in the Middle” effect, wherein models exhibit significantly weaker performance on relevant information located in the middle of long input contexts compared to information at the beginning or end. First identified by Liu et al., this effect demonstrates that even advanced long-context models, such as GPT-3.5-Turbo and Claude, exhibit strong primacy and recency biases in tasks like multi-document question answering and key-value retrieval, with performance forming a characteristic U-shaped curve \cite{liu2024lost}. This pattern has been further validated in recent benchmarks such as $\infty$Bench and RULER, where models consistently struggle not only with retrieval tasks but also with aggregation, reasoning, and variable tracking tasks that require deeper semantic understanding \cite{zhang2024bench, hsieh2024ruler}. Notably, even models that claim to support context lengths beyond 100K or 1M tokens often suffer substantial performance degradation well before reaching their stated limits, exposing fundamental architectural constraints in memory management, positional encoding, and attention mechanisms. Consequently, model hemorrhage reflects not only the computational strain of longer inputs but also a systemic decline in the ability to model long-range dependencies and integrate complex information.

Building on this, recent research \cite{long_context_rag} evaluated 20 large language models (LLMs) to assess the impact of context length variation on Retrieval-Augmented Generation (RAG) performance. While extending context length can offer performance gains, only a handful of state-of-the-art models maintain accuracy beyond 64,000 tokens. Even top-performing systems like GPT-4 Turbo \cite{achiam2023gpt} and Claude 3 show more than 20\% degradation when context length increases from 2K to 8K tokens. Moreover, very few models demonstrate reliable performance beyond 1 million tokens, underscoring inherent architectural limitations in current LLMs for extreme-context processing.

\section{AVENUES FOR FUTURE RESEARCH}\label{sec:challenges}

In this section, we outline several key research directions that we believe are critical for addressing "Model Hemorrhage" and improving the robustness of models.

\subsection{Model Doctor}  
\begin{figure}[h]
    \centering
    \includegraphics[width=0.6\linewidth]{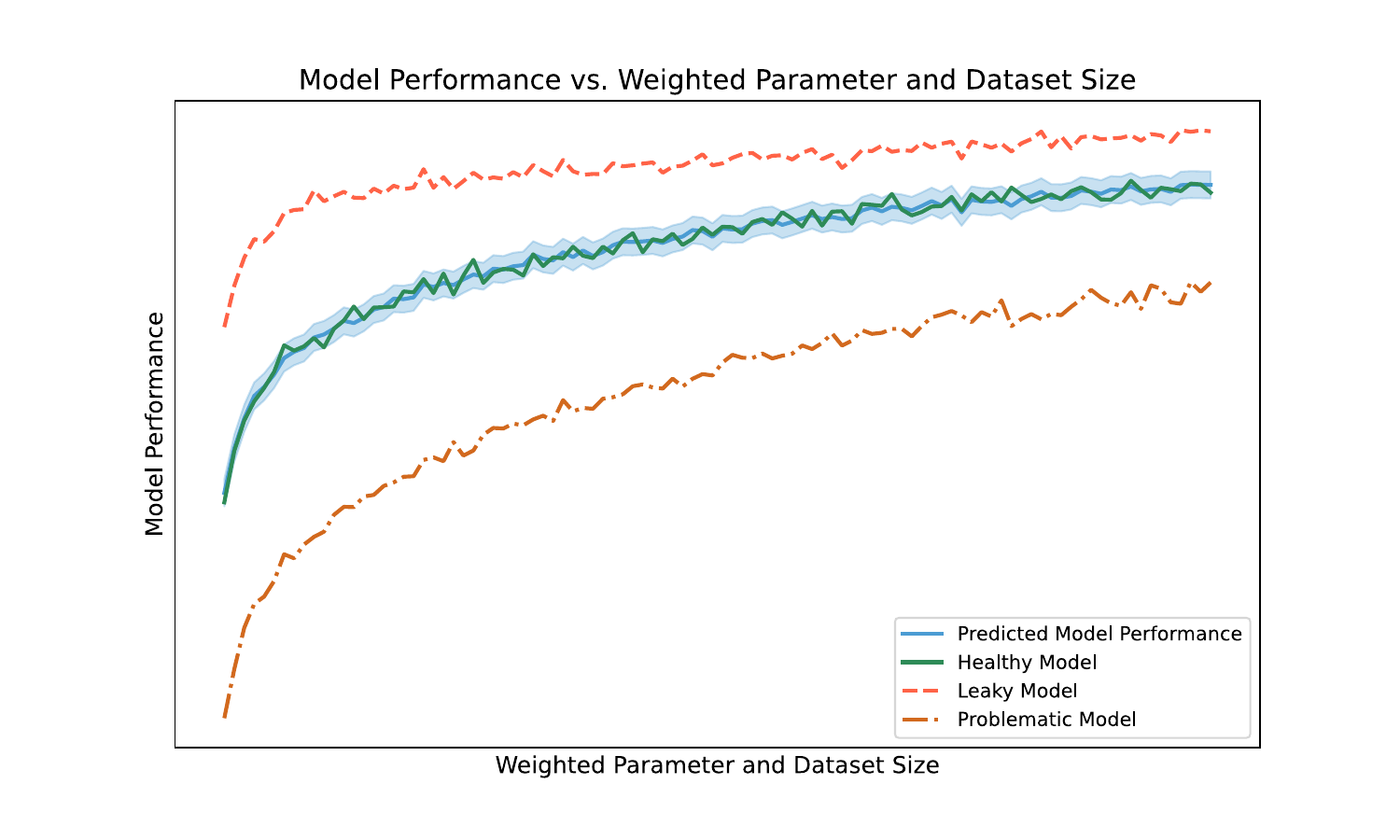}
    \caption{ The prediction and Real Metrics of Two Models.}
    \label{fig:health}
\end{figure}

To diagnose which models are "healthy," we need more advanced diagnostic frameworks. The concept of "Performance Law"\cite{performance_law_llms} offers a similar perspective by precisely predicting a model's performance on certain benchmarks based on its parameter size, architecture, and dataset scale. By comparing the actual results with the predicted scores, we can identify models that exhibit unexpected weaknesses, marking them as "unhealthy". Since model performance is generally positively correlated with the number of parameters and dataset size, it is feasible to train a "model doctor" by utilizing past performance data alongside model parameters. This approach allows us to predict the expected performance based on historical data. As shown in Figure \ref{fig:health}, when the actual performance of a model is significantly higher than the predicted value, it may indicate potential data leakage issues. Conversely, if the actual performance is much lower than expected, it could suggest underlying issues with the model's architecture or training methods. This diagnostic tool can provide valuable insights into the factors that may be affecting model performance, allowing researchers to address these issues and optimize model health.

The Hydra effect \cite{hydra_effect} and self-repair phenomena \cite{self_repair} in models are also worth attention. Perhaps some models exhibit significant self-healing after "Model Hemorrhage," which in turn enhances their scalability and robustness. These studies might offer theoretical support for structural pruning.

\subsection{New Architectures}  
Most current models predominantly rely on the Transformer architecture. However, emerging architectures, such as the “State Space Model (SSM)"-based Mamba model \cite{mamba}, may offer solutions to address robustness issues that are inherent to Transformer-based models. Exploring and evaluating these new architectures could pave the way for more robust and efficient models. Mamba shows significant advantages in inference speed and resource usage. Its integration with Transformer architectures in a proportional manner might resolve the inference challenges that Transformer models have faced. For example, Jamba \cite{jamba} integrates the attention layer and Mamba layer in a 1:7 ratio, combining Mamba's structure.

\subsection{Comprehensive Hemorrhage Testing and Evaluation Framework}  
Current research on robustness evaluation for large language models lacks a systematic framework. Our study aims to highlight this critical gap. Existing benchmark evaluations primarily focus on preserving model performance in fundamental tasks (e.g., perplexity, commonsense reasoning, and arithmetic problem-solving), failing to comprehensively assess capabilities in complex real-world scenarios \cite{tang2025lottery}. Furthermore, scant attention has been paid to evaluating a framework's suitability for iterative optimization and extensibility. Models demonstrating strong robustness under rigorous stress testing may prove more viable for in-depth academic investigation and deployment in performance-sensitive applications requiring exceptional stability.

%\section{IMPLICATIONS}\label{sec:conclusion}
\section{Conclusion}\label{sec:conclusion}
In this Perspective, we propose the "Model Hemorrhage" theoretical framework, systematically revealing multidimensional challenges in large language models (LLMs) through five critical dimensions: architectural redundancy, model compression, training-inference disparities, model extension, and data-related vulnerabilities. Our work comprehensively synthesizes precision degradation phenomena and limitations of existing mitigation approaches, establishing theoretical foundations for enhancing LLM robustness to improve reliability in real-world applications. This framework extends beyond conventional robustness paradigms constrained by data perturbations and adversarial attacks, emphasizing structural scalability and optimization-driven adaptations – critical considerations aligned with parameter expansion trends in modern LLMs.In this Perspective, we propose the "Model Hemorrhage" theoretical framework, systematically revealing multidimensional challenges in large language models (LLMs) through five critical dimensions: architectural redundancy, model compression, training-inference disparities, model extension, and data-related vulnerabilities. Our work comprehensively synthesizes precision degradation phenomena and limitations of existing mitigation approaches, establishing theoretical foundations for enhancing LLM robustness to improve reliability in real-world applications. This framework extends beyond conventional robustness paradigms constrained by data perturbations and adversarial attacks \cite{Croce2020, Wang2020, Hendrycks2020, Dong2021, Goel2021, Moradi2021, Wang2021, Wang2022, Malfa2022, chenunderstanding}, emphasizing structural scalability and optimization-driven adaptations – critical considerations aligned with parameter expansion trends in modern LLMs.

To validate the feasibility of the "Model Hemorrhage" framework, we conducted a series of experiments exploring the sensitivity of different models to optimization operations, including pruning, quantization, scaling, and decoding. Through systematic experimentation, we uncover pivotal trade-offs that redefine efficiency paradigms in language model deployment. For instance, in compression, while structured pruning accelerates inference, it incurs severe accuracy loss (above 50\% sparsity triggers catastrophic failure) outperformed by unstructured and quantization methods. Quantization exhibits a "safe compression zone" (above 2-bit precision), beyond which performance degrades nonlinearly—yet larger models retain superior robustness under low-bit settings compared to smaller counterparts. With regard to Decoding Methods, deterministic decoding approaches exhibited superior and more stable performance compared to stochastic ones.

As the parameters of large-scale language models (LLMs) continue to expand, the marginal benefits of increasing model size diminish, while resource consumption escalates. We call for urgent attention to this phenomenon. Future work should focus on refining model health diagnostic methods, exploring new architectures, and developing a systematic robustness evaluation framework to address the challenges posed by large-scale language models in dynamic environments.

\backmatter

%\section*{Supplementary information}

%If your article has accompanying supplementary file/s please state so here. 

%Authors reporting data from electrophoretic gels and blots should supply the full unprocessed scans for key as part of their Supplementary information. This may be requested by the editorial team/s if it is missing. Please refer to Journal-level guidance for any specific requirements.

%\section*{Acknowledgements}

%Acknowledgements are not compulsory. Where included they should be brief. Grant or contribution numbers may be acknowledged. Please refer to Journal-level guidance for any specific requirements.

%\section*{Author contributions}

\section*{Competing interests}
The authors declare no competing interests.
%Some journals require declarations to be submitted in a standardised format. Please check the Instructions for Authors of the journal to which you are submitting to see if you need to complete this section. If yes, your manuscript must contain the following sections under the heading `Declarations':

%\begin{itemize}
%\item Conflict of interest/Competing interests (check journal-specific guidelines for %which heading to use)
%\item Ethics approval and consent to participate
%\item Consent for publication
%\item Data availability 
%\end{itemize}

%\noindent
%If any of the sections are not relevant to your manuscript, please include the heading and write `Not applicable' for that section. 

%\bigskip

%\bigskip\noindent
%Springer journals and proceedings: \url{https://www.springer.com/gp/editorial-policies}

%\bigskip\noindent
%Nature Portfolio journals: \url{https://www.nature.com/nature-research/editorial-policies}

%\bigskip\noindent
%\textit{Scientific Reports}: \url{https://www.nature.com/srep/journal-policies/editorial-policies}

%\bigskip\noindent
%BMC journals: \url{https://www.biomedcentral.com/getpublished/editorial-policies}
%\end{flushleft}

\section*{Additional information}
Correspondence and requests for materials should be addressed to Zuchao Li.

\begin{appendices}

\section{Performance and Robustness Under Model Hemorrhage}\label{sec:experiments}
%An appendix contains supplementary information that is not an essential part of the text itself but which may be helpful in providing a more comprehensive understanding of the research problem or it is information that is too cumbersome to be included in the body of the paper.

In this section, we will conduct a performance and robustness comparison of models with similar parameter sizes under various typical model hemorrhage scenarios. This comparison will be based on publicly available performance data, relevant research, and our own experimental results. In addition to the factors mentioned in Section 3, we will also investigate how data disturbances, adversarial attacks, and other data-related environments contribute to model hemorrhage.

\subsection{Sensitivity to Structured Pruning}\label{sec:structured_pruning_exp}
\begin{figure*}[ht]
    \centering
    \includegraphics[width=0.8\linewidth]{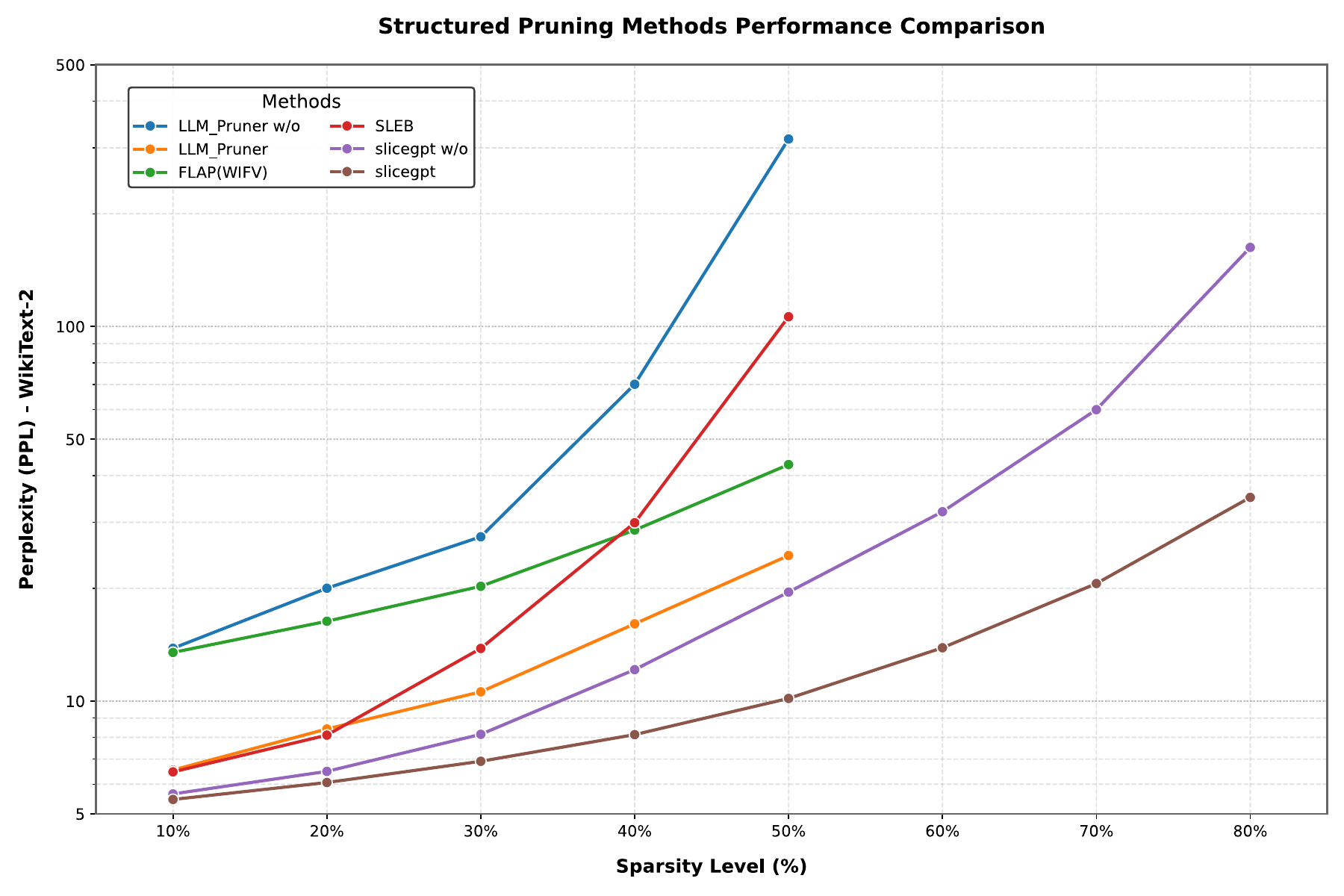}
    \caption{This image is a performance comparison of different structured pruning methods applied to LLaMA2-7b, based on the WikiText2-PPL (Perplexity) metric, across varying sparsity levels. The chart highlights the impact of pruning on model performance, with lower PPL values indicating better performance.}
    \label{fig:structured_ppl}
\end{figure*}

In this study, we focused on reproducing several typical structured pruning methods, using LLaMA2-7b as the experimental model to ensure compatibility with different pruning techniques. The primary goal was to evaluate the performance of these methods under varying compression rates, measured by WikiText2-PPL (perplexity), where lower PPL indicates better model performance. We compared how structured pruning affects the WikiText2-PPL at different sparsity levels, aiming to better understand the trade-off between model accuracy and pruning-induced speedup (Table \ref{tab:structured_ppl}).
The experimental results indicate that structured pruning leads to model collapse when sparsity reaches around 50\% (Fig \ref{fig:structured_ppl}). In some cases, the model can be fine-tuned post-pruning to mitigate the damage, thereby raising the threshold at which collapse occurs. However, regardless of fine-tuning, structured pruning consistently results in immediate Model Hemorrhage with each reduction in the model’s structure.

While the main advantage of structured pruning lies in inference speedup, as highlighted in several recent papers, practical observations reveal that the trade-off between accuracy and speed is not always favorable. When sparsity exceeds 50\%, the model generally collapses, providing less than a twofold speedup. This result demonstrates that, after a certain point, the benefits in terms of speed are outweighed by significant losses in model performance, challenging the practicality of structured pruning for achieving high compression rates without compromising model effectiveness.

\begin{table}[ht]
\centering
\caption{Wikitext2-PPL results for structured pruning (w/o: without finetuning).}
\label{tab:structured_ppl}
\small
\begin{tabular}{|l|c|c|c|c|c|c|c|c|}
\hline
Method & 0.1 & 0.2 & 0.3 & 0.4 & 0.5 & 0.6 & 0.7 & 0.8 \\
\hline
LLM\_Pruner w/o \cite{llm_pruner}& 13.8316 & 20.0072 & 27.4537 & 70.1054 & 316.6549 & - & - & -\\
LLM\_Pruner \cite{llm_pruner}& 6.5463 & 8.4221 & 10.5843 & 16.0763 & 24.4655 & - & - & -\\
FLAP(WIFV) \cite{flap_pruning}& 13.4835 & 16.3325 & 20.2505 & 28.6018 & 42.7961 & - & - & -\\
SLEB \cite{SLEB}& 6.4700 & 8.1100 & 13.8200 & 29.9300 & 106.1900 & - & - & -\\
slicegpt w/o \cite{slice_gpt}& 5.6500 & 6.4900 & 8.1500 & 12.1300 & 19.5400 & 32.0300 & 59.9800 & 162.65\\
slicegpt \cite{slice_gpt}& \textbf{5.4610} & \textbf{6.0607} & \textbf{6.9047} & \textbf{8.1353} & \textbf{10.1622} & \textbf{13.8773} & \textbf{20.5934} & \textbf{34.9808}\\
\hline
\end{tabular}
\end{table}

Similarly, We employed SliceGPT~\cite{slice_gpt} to prune both the LLaMA family (LLaMA-2 \cite{touvron2023llama} and LLaMA-3 \cite{llama3herd2024}) and the OPT family \cite{zhang2022opt}, comparing the impact of pruning on their performance in terms of perplexity (PPL) on the WikiText-2 dataset. 

\begin{figure*}[ht]
    \centering
    \includegraphics[width=0.8\linewidth]{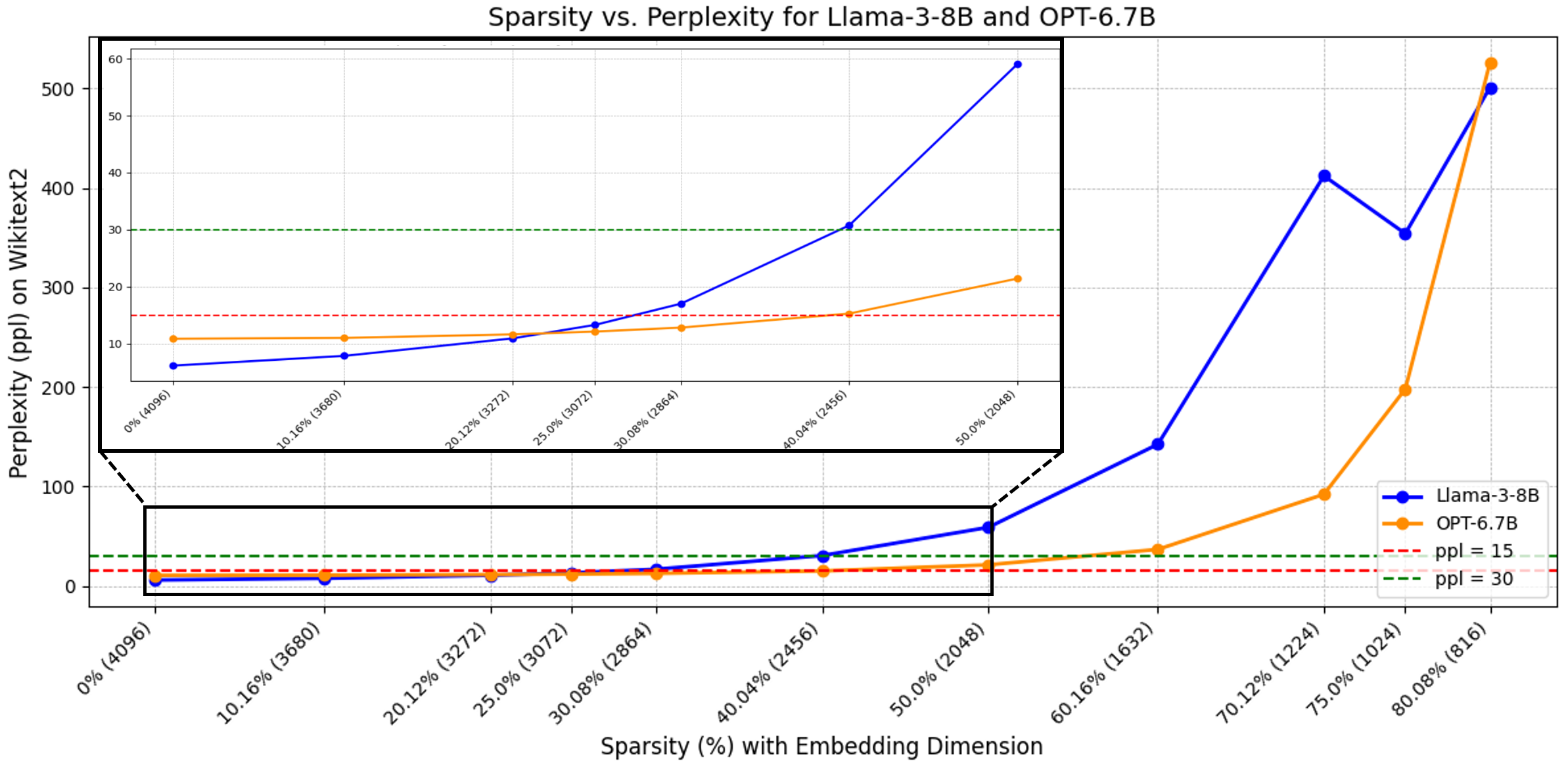}
    \caption{Perplexity performance of LLaMA-3-8B and OPT-6.7B under different pruning ratios.}
    \label{fig:slice}
\end{figure*}

From Fig.~\ref{fig:slice}, it can be observed that when LLaMA-3-8B is pruned beyond 50\%, there is a significant increase in PPL, indicating performance degradation. In contrast, OPT-6.7B exhibits notable PPL loss only after pruning reaches 70\%. 

\subsection{Experimental Evaluation of Unstructured Pruning Methods}\label{sec:unstructured_pruning_exp}
\begin{figure*}[ht]
    \centering
    \includegraphics[width=0.8\linewidth]{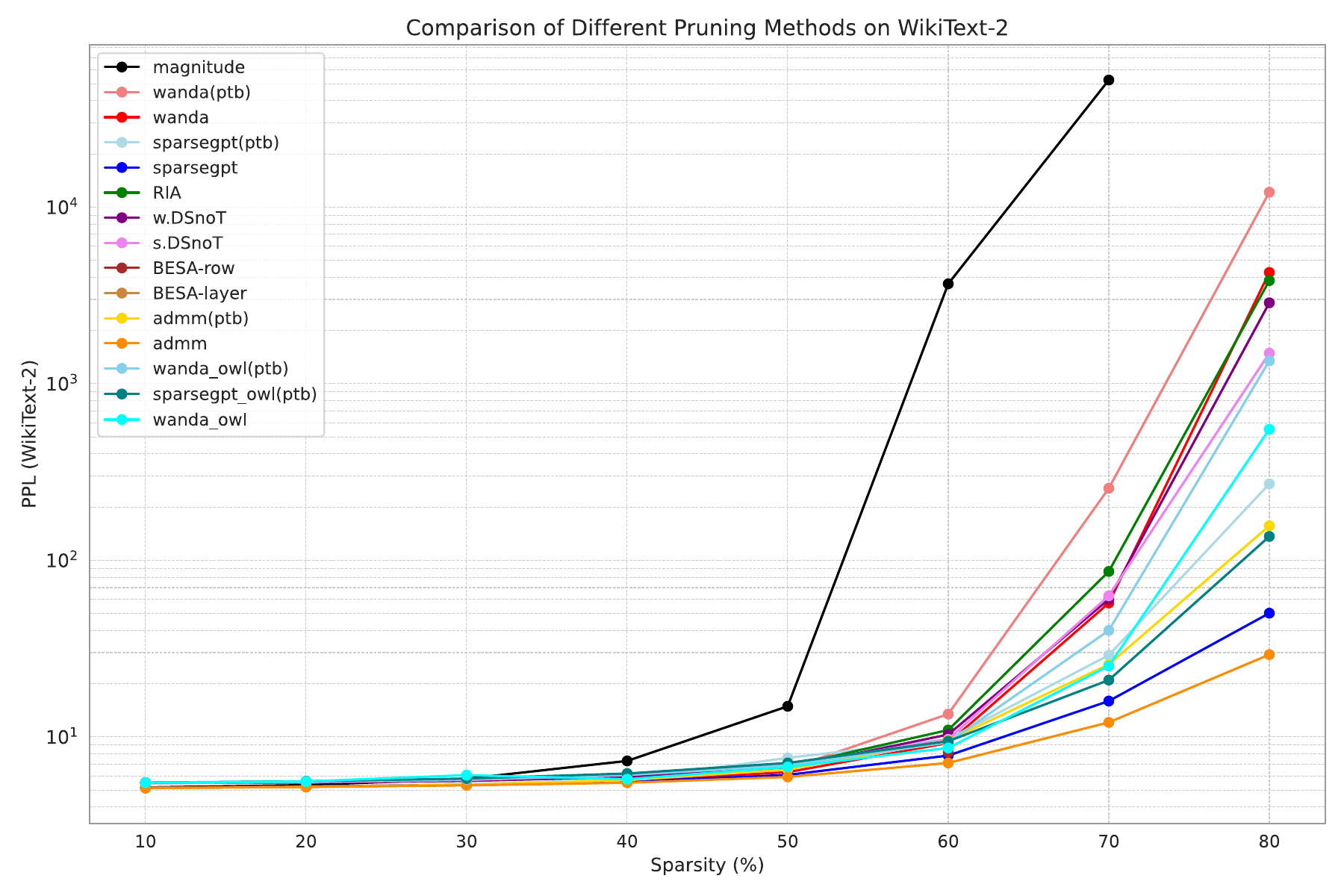}
    \caption{ WikiText2-PPL Trend with Unstructured Pruning Methods Across Varying Compression Rates (10\% - 80\%).This image demonstrates how the WikiText2-PPL (Perplexity) of LLaMA2-7b changes as the pruning compression rate increases from 10\% to 80\%. It should be noted that methods using the "PTB" calibration dataset (indicated with "(ptb)" after the method name) are included as part of an ablation study. As the pruning rate rises, many methods experience significant performance loss, and some even result in model collapse at very high compression rates.}
    \label{fig:unstructured_ppl}
\end{figure*}

\begin{figure*}[!htbp]
    \centering
    \includegraphics[width=0.8\linewidth]{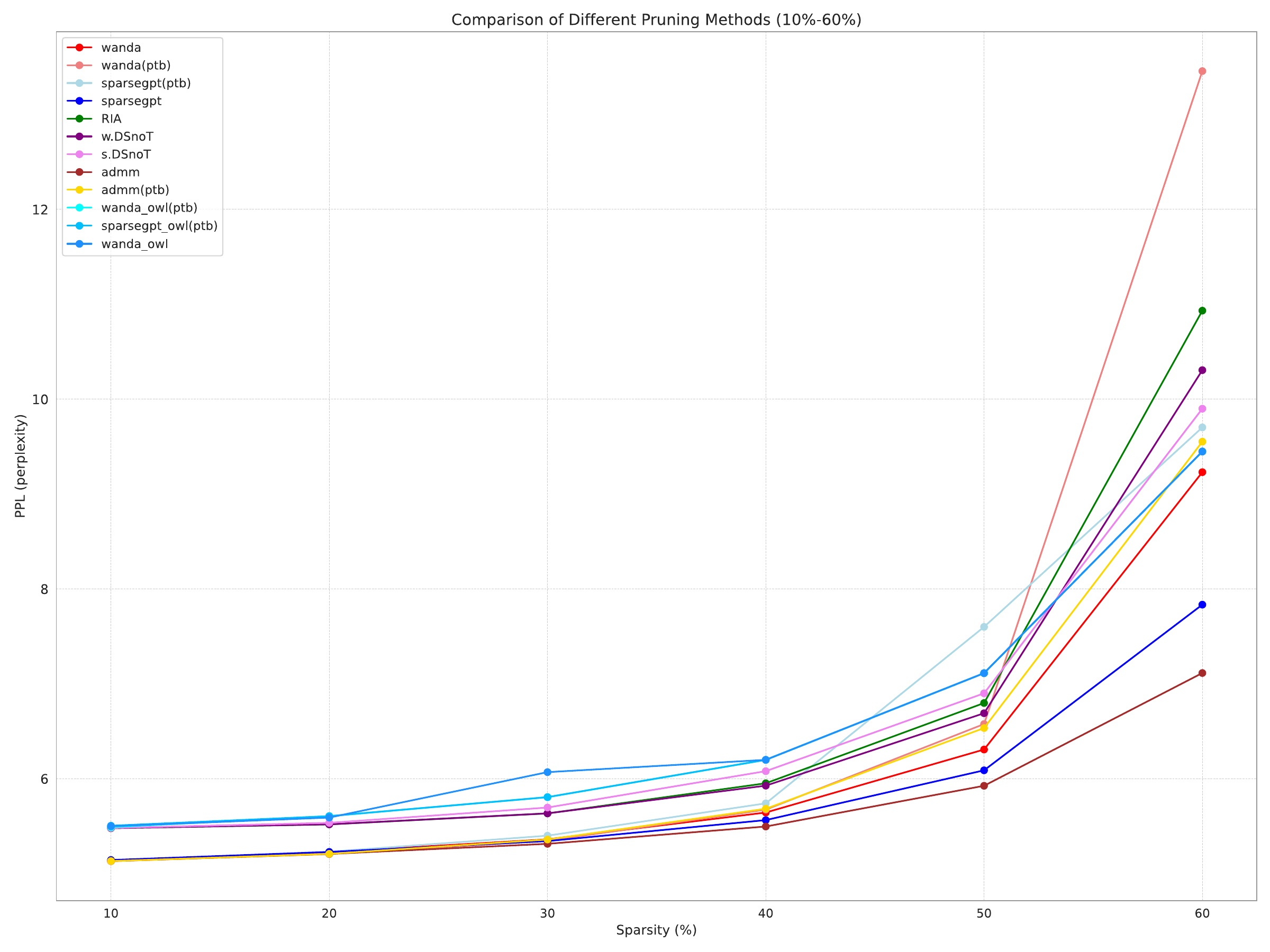}
    \caption{WikiText2-PPL Trend with Unstructured Pruning Methods at Compression Rates Below 60\% for LLaMA2-7b. This figure contrasts various unstructured pruning methods, with Wanda and SparseGPT in red and blue, respectively. As shown, at compression rates below 60\%, these two methods exhibit less performance degradation compared to most other approaches. It should be noted that methods using the "PTB" calibration dataset (indicated with "(ptb)" after the method name) are included as part of an ablation study. ADMM, Wanda and SparseGPT offer more stable and robust performance compared to other pruning methods.}
    \label{fig:unstructured_ppl_60}
\end{figure*}

In the past two years, substantial progress has been made in the field of unstructured pruning, with methods such as SparseGPT and Wanda frequently serving as baseline or foundational approaches. Notably, many studies highlight performance under high compression rates (>=70\%) as a major selling point. However, upon closer inspection, it becomes clear that models pruned at these high compression rates experience significant performance degradation, with some even collapsing entirely (Fig \ref{fig:unstructured_ppl}). This highlights a critical flaw in using such high-compression scenarios as a measure of success. In contrast, when the compression rate is kept below 60\%, methods like Wanda and SparseGPT consistently demonstrate a more minimal impact on model performance compared to most other approaches (Fig \ref{fig:unstructured_ppl_60}). This suggests that these methods are more robust and effective in maintaining model quality at lower sparsity levels, making them more practical for real-world applications where model performance is paramount. Specific experimental data on unstructured pruning is referenced in Table \ref{tab:unmistructured_ppl}. Additionally, we conducted ablation experiments on the Calibration dataset, and the alignment results on the PTB dataset can be found in Table \ref{tab:unmistructured_ppl_ptb}.


\subsection{Comparison of Unstructured Pruning and Semi-structured Pruning Methods}\label{sec:semi-structured_pruning_exp}

\begin{figure*}[!htbp]
    \centering
    \includegraphics[width=0.6\linewidth]{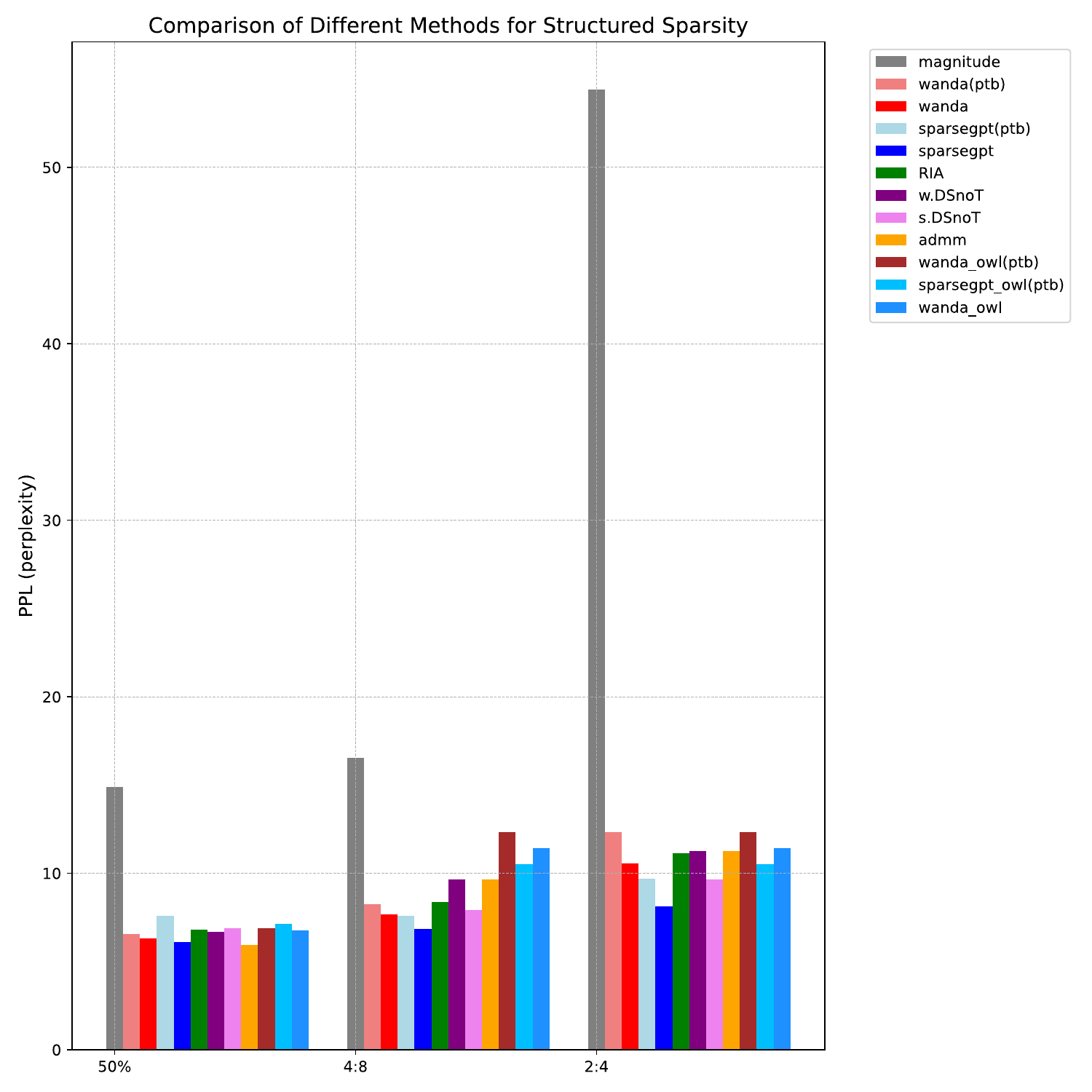}
    \caption{Comparison of the performance of unstructured and structured pruning methods on WikiText2 PPL at 50\% sparsity. The figure compares the perplexity (PPL) of various pruning methods for a model pruned to 50\% sparsity. The methods include unstructured pruning (50\%), and two types of structured pruning: 4:8 and 2:4, which indicate that 4 out of every 8 weights or 2 out of every 4 weights are pruned, respectively.}
    \label{fig:semistructured_ppl}
\end{figure*}

Following the experiments on unstructured pruning in Section \ref{sec:unstructured_pruning_exp}, we also conducted experiments on Semi-structured pruning using similar methods. The pruning sparsity was set to 50\%, and we specifically employed the 4:8 and 2:4 configurations, which represent pruning four out of eight weights, or two out of four weights, respectively. When comparing the results of unstructured pruning and Semi-structured pruning, we observed that Semi-structured pruning using the same method resulted in a higher accuracy loss than unstructured pruning  (Fig \ref{fig:semistructured_ppl}). Based on these findings, unstructured pruning is preferred under the same conditions due to its better performance. Detailed experimental data for semi-structured pruning at 50\% sparsity is provided in Table \ref{tab:unmistructured_ppl}. Additionally, we performed ablation experiments using the Calibration dataset, with alignment results on the PTB dataset shown in Table \ref{tab:unmistructured_ppl_ptb}.

\begin{table}[ht]
\centering
\small
\caption{Detailed Experimental Data on WikiText2-PPL for Different Unstructured and Semi-structured Methods (The calibration dataset is WikiText2). The best-performing methods are marked in bold, while the second-best methods are underlined.}
\label{tab:unmistructured_ppl}
\setlength{\tabcolsep}{3pt}
\begin{tabular}{|l|c|c|c|c|c|c|c|c|c|c|}
\hline
\textbf{Method} & 0.1 & 0.2 & 0.3 & 0.4 & 0.5 & 0.6 & 0.7 & 0.8 & 2:4(0.5) & 4:8(0.5) \\
\hline
Magnitude & 5.1737 & 5.3292 & 5.7920 & 7.3069 & 14.8954 & 3676.1436 & 52422.6016 & nan & 54.3870 & 16.5288 \\
Wanda \cite{wanda}& \underline{5.1352} & \underline{5.2269} & 5.3609 & 5.6434 & 6.3075 & 9.2307 & 57.2495 & 4262.4902 & 10.5414 & 7.6748 \\
SparseGPT \cite{sparsegpt}& 5.1428 & 5.2277 & \underline{5.3423} & \underline{5.5635} & \underline{6.0882} & \underline{7.8341} & \underline{15.9474} & \underline{50.1593} & \underline{8.1100} & \textbf{6.8299} \\
RIA \cite{RIA}& 5.4777 & 5.5181 & 5.6338 & 5.9517 & 6.7974 & 10.9329 & 86.4084 & 3832.7690 & 11.1520 & 8.3561 \\
w.DSnoT \cite{dsnoprune}& 5.4789 & 5.5190 & 5.6344 & 5.9264 & 6.6898 & 10.3059 & 60.1499 & 2870.5852 & 11.2686 & 9.6594 \\
s.DSnoT \cite{dsnoprune}& 5.4805 & 5.5347 & 5.6958 & 6.0794 & 6.8997 & 9.6878 & 62.8766 & 1487.4147 & 9.6594 & 7.9369 \\
ADMM \cite{admm}& \textbf{5.1315} & \textbf{5.2076} & \textbf{5.3134} & \textbf{5.4959} & \textbf{5.9252} & \textbf{7.1137} & \textbf{12.0747} & \textbf{29.2164} & \textbf{7.5892} & \underline{7.5892} \\
Wanda\_owl \cite{yin2024outlier}& 5.4989 & 5.5901 & 6.0690 & 5.7550 & 6.7490 & 8.6492 & 25.1889 & 551.7274 & 11.4216 & - \\
BESA-row \cite{BESA}& - & - & - & - & 6.0057 & - & 8.1769 & - & - & - \\
BESA-layer \cite{BESA}& - & - & - & - & 5.9884 & - & 9.0709 & - & - & - \\
\hline
\end{tabular}
\end{table}
\begin{table}[!htbp]
\centering
\small
\caption{Detailed Experimental Data on WikiText2-PPL for Different Unstructured and Semi-structured Methods (The calibration dataset is PTB). The best-performing methods are marked in bold, while the second-best methods are underlined.}
\label{tab:unmistructured_ppl_ptb}
\begin{tabular}{|l|c|c|c|c|c|c|c|c|c|c|}
\hline
\textbf{Method} & 0.1 & 0.2 & 0.3 & 0.4 & 0.5 & 0.6 & 0.7 & 0.8 & 2:4(0.5) & 4:8(0.5) \\
\hline
wanda \cite{wanda}& \underline{5.1313} & \textbf{5.205} & \textbf{5.3455} & \textbf{5.6736} & \underline{6.5758} & 13.4577 & 255.5912 & 12142.1484 & 12.5204 & \underline{8.2569} \\
sparsegpt \cite{sparsegpt}& 5.135 & 5.2305 & 5.3994 & 5.74 & 7.5999 & 9.7025 & 29.0151 & 270.8205 & \textbf{9.7025} & \textbf{7.5999} \\
wanda\_owl \cite{yin2024outlier}& 5.4885 & 5.5719 & 5.7487 & 6.0911 & 6.8722 & \textbf{9.3082} & 40.0371 & 1346.3525 & 12.3241 & - \\
sparsegpt\_owl \cite{yin2024outlier}& 5.5043 & 5.6054 & 5.8055 & 6.1986 & 7.1123 & \underline{9.4483} & \textbf{20.963} & \textbf{136.4502} & 10.5258 & - \\
admm \cite{admm}& \textbf{5.1292} & \underline{5.2079} & \underline{5.3587} & \underline{5.6827} & \textbf{6.5338} & 9.5521 & \underline{25.6936} & \underline{156.7145} & \underline{10.0416} & 10.0416 \\
\hline
\end{tabular}
\end{table}

\subsection{Sensitivity to Quantization}\label{sec:quantization_exp}

\begin{figure}[h]
    \centering
    \includegraphics[width=1\linewidth]{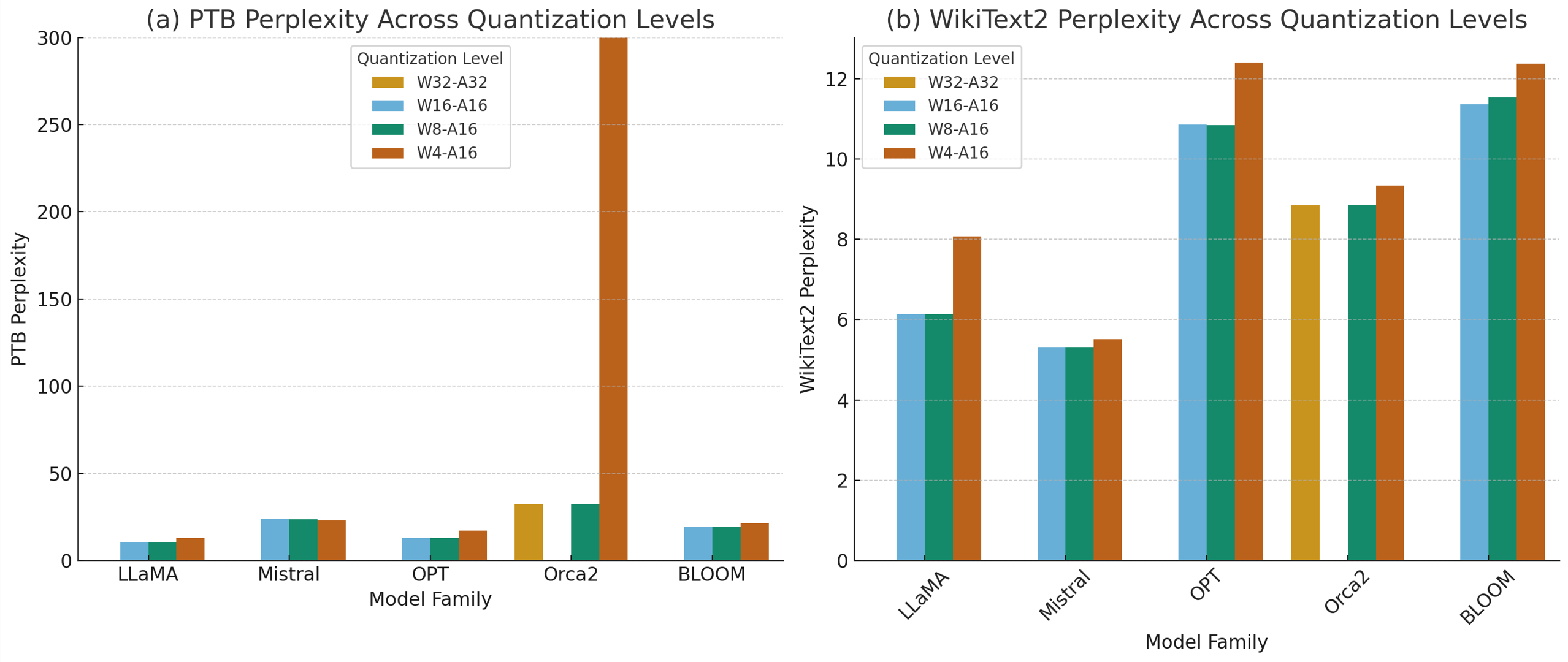}
    \caption{Perplexity Across Quantization Levels. (a) PTB. (b) Wikitext2.}
    \label{fig:gptq}
\end{figure}

We conducted weight-only quantization experiments on base models of approximately 7B parameters using GPTQ \cite{gptq} with 8-bit and 4-bit precision. The models evaluated include LLaMA-3-8B \cite{llama3herd2024}, Mistral-7B \cite{mistral7b}, OPT-6.7B \cite{zhang2022opt}, Orca2-7B \cite{orca2}, and BLOOM-7.1B \cite{bloom}. The results, presented in Figure~\ref{fig:gptq}, indicate that most models exhibit stable performance after both 8-bit and 4-bit quantization. Mistral-7B demonstrated the smallest change in perplexity, while LLaMA-3-8B, OPT-6.7B, and BLOOM-7.1B experienced minor performance degradation at 4-bit quantization but remained relatively robust overall. In contrast, Orca2-7B displayed a significant deterioration in performance, highlighting its weaker resistance to weight-only quantization.

\definecolor{DeepBlue}{RGB}{0,51,102}

\begin{table*}[htbp]
\centering
\scriptsize
\setlength{\tabcolsep}{4pt}
\caption{Comprehensive evaluation of quantized models including perplexities (PPL), inference speed, parameter sizes, and downstream task performance.}
\label{tab:full-results}
\begin{tabular}{l l c c c c c c c c}
\toprule
\multirow{2}{*}{\textbf{Main Model}} & \multirow{2}{*}{\textbf{Indicator}} 
& \multicolumn{8}{c}{\textbf{Quantization (bits)}} \\
\cmidrule(lr){3-10}
& & \textbf{1} & \textbf{2} & \textbf{3} & \textbf{4} & \textbf{5} & \textbf{6} & \textbf{8} & \textbf{16}\\
\midrule

%% ============ Qwen2.5-0.5B ============
\multirow{7}{*}{\textbf{Qwen2.5-0.5B}}
& \cellcolor{DeepBlue}\textbf{\textcolor{white}{Quant Scheme}} 
  & \cellcolor{DeepBlue}\textcolor{white}{-} 
  & \cellcolor{DeepBlue}\textcolor{white}{Q2\_K} 
  & \cellcolor{DeepBlue}\textcolor{white}{Q3\_K\_M} 
  & \cellcolor{DeepBlue}\textcolor{white}{Q4\_K\_M} 
  & \cellcolor{DeepBlue}\textcolor{white}{Q5\_K\_M} 
  & \cellcolor{DeepBlue}\textcolor{white}{Q6\_K} 
  & \cellcolor{DeepBlue}\textcolor{white}{Q8\_0} 
  & \cellcolor{DeepBlue}\textcolor{white}{FP16} \\
& \textbf{PPL}
  & - & \makecell{14.8206\\[-0.4em]\textcolor{orange}{\tiny $\pm0.1111$}} & \makecell{14.2463\\[-0.4em]\textcolor{orange}{\tiny $\pm0.1062$}} & \makecell{13.9797\\[-0.4em]\textcolor{orange}{\tiny $\pm0.1039$}} & \makecell{13.9324\\[-0.4em]\textcolor{orange}{\tiny $\pm0.1038$}} & \makecell{13.7029\\[-0.4em]\textcolor{orange}{\tiny $\pm0.1018$}} & \makecell{13.7042\\[-0.4em]\textcolor{orange}{\tiny $\pm0.1018$}} & \makecell{13.6439\\[-0.4em]\textcolor{orange}{\tiny $\pm0.1012$}} \\
& \textbf{Tokens/s}
  & - & 5593.6 & 5343.97 & 5370.62 & 5385.33 & 5427.74 & 5465.18 & 4266.22 \\
& \textbf{Param(GB)}
  & - & 0.41 & 0.42 & 0.48 & 0.52 & 0.63 & 0.66 & 1.27 \\
& \textbf{ARC-C}
  & - & 0.43 & 0.35 & 0.39 & 0.39 & 0.37 & 0.38 & 0.38 \\
& \textbf{ARC-E}
  & - & 0.62 & 0.45 & 0.56 & 0.59 & 0.59 & 0.61 & 0.61 \\
& \textbf{MMLU}
  & - & 0.231 & 0.236 & 0.235 & 0.231 & 0.231 & 0.231 & 0.231 \\
\midrule

%% ============ Qwen2.5-1.5B ============
\multirow{7}{*}{\textbf{Qwen2.5-1.5B}}
& \cellcolor{DeepBlue}\textbf{\textcolor{white}{Quant Scheme}} 
  & \cellcolor{DeepBlue}\textcolor{white}{-} 
  & \cellcolor{DeepBlue}\textcolor{white}{Q2\_K} 
  & \cellcolor{DeepBlue}\textcolor{white}{Q3\_K\_M} 
  & \cellcolor{DeepBlue}\textcolor{white}{Q4\_K\_M} 
  & \cellcolor{DeepBlue}\textcolor{white}{Q5\_K\_M} 
  & \cellcolor{DeepBlue}\textcolor{white}{Q6\_K} 
  & \cellcolor{DeepBlue}\textcolor{white}{Q8\_0} 
  & \cellcolor{DeepBlue}\textcolor{white}{FP16} \\
& \textbf{PPL}
  & - & \makecell{14.6341\\[-0.4em]\textcolor{orange}{\tiny $\pm0.1034$}} & \makecell{10.6066\\[-0.4em]\textcolor{orange}{\tiny $\pm0.0733$}} & \makecell{10.0107\\[-0.4em]\textcolor{orange}{\tiny $\pm0.0684$}} & \makecell{9.7016\\[-0.4em]\textcolor{orange}{\tiny $\pm0.0659$}} & \makecell{9.6790\\[-0.4em]\textcolor{orange}{\tiny $\pm0.0658$}} & \makecell{9.6617\\[-0.4em]\textcolor{orange}{\tiny $\pm0.0657$}} & \makecell{9.6555\\[-0.4em]\textcolor{orange}{\tiny $\pm0.0657$}} \\
& \textbf{Tokens/s}
  & - & 3814.78 & 3630.17 & 3843.72 & 3494.1 & 3543.53 & 3265.56 & 2133.91 \\
& \textbf{Param(GB)}
  & - & 0.74 & 0.90 & 1.12 & 1.29 & 1.46 & 1.89 & 3.56 \\
& \textbf{ARC-C}
  & - & 0.47 & 0.68 & 0.75 & 0.75 & 0.74 & 0.75 & 0.77 \\
& \textbf{ARC-E}
  & - & 0.56 & 0.86 & 0.90 & 0.91 & 0.91 & 0.90 & 0.90 \\
& \textbf{MMLU}
  & - & 0.315 & 0.579 & 0.586 & 0.603 & 0.603 & 0.598 & 0.601 \\
\midrule

%% ============ Qwen2.5-3B ============
\multirow{7}{*}{\textbf{Qwen2.5-3B}}
& \cellcolor{DeepBlue}\textbf{\textcolor{white}{Quant Scheme}} 
  & \cellcolor{DeepBlue}\textcolor{white}{-} 
  & \cellcolor{DeepBlue}\textcolor{white}{Q2\_K} 
  & \cellcolor{DeepBlue}\textcolor{white}{Q3\_K\_M} 
  & \cellcolor{DeepBlue}\textcolor{white}{Q4\_K\_M} 
  & \cellcolor{DeepBlue}\textcolor{white}{Q5\_K\_M} 
  & \cellcolor{DeepBlue}\textcolor{white}{Q6\_K} 
  & \cellcolor{DeepBlue}\textcolor{white}{Q8\_0} 
  & \cellcolor{DeepBlue}\textcolor{white}{FP16} \\
& \textbf{PPL}
  & - & \makecell{12.3285\\[-0.4em]\textcolor{orange}{\tiny $\pm0.0904$}} & \makecell{9.7013\\[-0.4em]\textcolor{orange}{\tiny $\pm0.0684$}} & \makecell{9.2818\\[-0.4em]\textcolor{orange}{\tiny $\pm0.0649$}} & \makecell{9.2155\\[-0.4em]\textcolor{orange}{\tiny $\pm0.0646$}} & \makecell{9.1428\\[-0.4em]\textcolor{orange}{\tiny $\pm0.0641$}} & \makecell{9.1195\\[-0.4em]\textcolor{orange}{\tiny $\pm0.0639$}} & - \\
& \textbf{Tokens/s}
  & - & 2397.99 & 2166.52 & 2117.6 & 2005.48 & 1669.51 & 1427.8 & - \\
& \textbf{Param(GB)}
  & - & 1.38 & 1.72 & 2.10 & 2.44 & 2.79 & 3.62 & - \\
& \textbf{ARC-C}
  & - & 0.67 & 0.79 & 0.79 & 0.86 & 0.84 & 0.85 & - \\
& \textbf{ARC-E}
  & - & 0.80 & 0.92 & 0.93 & 0.92 & 0.92 & 0.92 & - \\
& \textbf{MMLU}
  & - & 0.500 & 0.604 & 0.649 & 0.657 & 0.656 & 0.664 & - \\
\midrule
%% ============ Qwen2.5-7B ============
\multirow{7}{*}{\textbf{Qwen2.5-7B}}
& \cellcolor{DeepBlue}\textbf{\textcolor{white}{Quant Scheme}} 
  & \cellcolor{DeepBlue}\textcolor{white}{-} 
  & \cellcolor{DeepBlue}\textcolor{white}{IQ2\_M} 
  & \cellcolor{DeepBlue}\textcolor{white}{Q3\_K\_M} 
  & \cellcolor{DeepBlue}\textcolor{white}{Q4\_K\_M} 
  & \cellcolor{DeepBlue}\textcolor{white}{Q5\_K\_M} 
  & \cellcolor{DeepBlue}\textcolor{white}{Q6\_K} 
  & \cellcolor{DeepBlue}\textcolor{white}{Q8\_0} 
  & \cellcolor{DeepBlue}\textcolor{white}{FP16} \\
& \textbf{PPL}
  & - & \makecell{9.2085\\[-0.4em]\textcolor{orange}{\tiny $\pm0.0627$}} & \makecell{8.1920\\[-0.4em]\textcolor{orange}{\tiny $\pm0.0550$}} & \makecell{7.9982\\[-0.4em]\textcolor{orange}{\tiny $\pm0.0533$}} & \makecell{7.9628\\[-0.4em]\textcolor{orange}{\tiny $\pm0.0529$}} & \makecell{7.9505\\[-0.4em]\textcolor{orange}{\tiny $\pm0.0528$}} & \makecell{7.9493\\[-0.4em]\textcolor{orange}{\tiny $\pm0.0528$}} & - \\
& \textbf{Tokens/s}
  & - & 1439.49 & 952.80 & 1014.46 & 727.07 & 759.87 & 520.07 & - \\
& \textbf{Param(GB)}
  & - & 2.78 & 3.81 & 4.68 & 5.44 & 6.25 & 8.10 & - \\
& \textbf{ARC-C}
  & - & 0.83 & 0.87 & 0.87 & 0.86 & 0.86 & 0.87 & - \\
& \textbf{ARC-E}
  & - & 0.94 & 0.96 & 0.96 & 0.96 & 0.96 & 0.95 & - \\
& \textbf{MMLU}
  & - & 0.482 & 0.688 & 0.707 & 0.675 & 0.702 & 0.683 & - \\
\midrule
%% ============ Qwen2.5-14B ============
\multirow{7}{*}{\textbf{Qwen2.5-14B}}
& \cellcolor{DeepBlue}\textbf{\textcolor{white}{Quant Scheme}} 
  & \cellcolor{DeepBlue}\textcolor{white}{-} 
  & \cellcolor{DeepBlue}\textcolor{white}{Q2\_K} 
  & \cellcolor{DeepBlue}\textcolor{white}{Q3\_K\_M} 
  & \cellcolor{DeepBlue}\textcolor{white}{Q4\_K\_M} 
  & \cellcolor{DeepBlue}\textcolor{white}{Q5\_K\_M} 
  & \cellcolor{DeepBlue}\textcolor{white}{Q6\_K} 
  & \cellcolor{DeepBlue}\textcolor{white}{Q8\_0} 
  & \cellcolor{DeepBlue}\textcolor{white}{FP16} \\
& \textbf{PPL}
  & - & \makecell{7.7406\\[-0.4em]\textcolor{orange}{\tiny $\pm0.0515$}} & \makecell{6.9516\\[-0.4em]\textcolor{orange}{\tiny $\pm0.0446$}} & \makecell{6.8339\\[-0.4em]\textcolor{orange}{\tiny $\pm0.0437$}} & \makecell{6.8031\\[-0.4em]\textcolor{orange}{\tiny $\pm0.0435$}} & \makecell{6.7826\\[-0.4em]\textcolor{orange}{\tiny $\pm0.0434$}} & \makecell{6.7821\\[-0.4em]\textcolor{orange}{\tiny $\pm0.0433$}} & - \\
& \textbf{Tokens/s}
  & - & 320.36 & 311.41 & 303.88 & 314.02 & 257.84 & 185.52 & - \\
& \textbf{Param(GB)}
  & - & 5.77 & 7.34 & 8.99 & 10.50 & 12.10 & 15.70 & - \\
& \textbf{ARC-C}
  & - & 0.87 & 0.90 & 0.91 & 0.92 & 0.93 & 0.92 & - \\
& \textbf{ARC-E}
  & - & 0.98 & 1.00 & 0.99 & 1.00 & 1.00 & 1.00 & - \\
& \textbf{MMLU}
  & - & 0.731 & 0.766 & 0.787 & 0.777 & 0.772 & 0.784 & - \\
\midrule
%% ============ Qwen2.5-32B ============
\multirow{7}{*}{\textbf{Qwen2.5-32B}}
& \cellcolor{DeepBlue}\textbf{\textcolor{white}{Quant Scheme}} 
  & \cellcolor{DeepBlue}\textcolor{white}{-} 
  & \cellcolor{DeepBlue}\textcolor{white}{Q2\_K} 
  & \cellcolor{DeepBlue}\textcolor{white}{Q3\_K\_M} 
  & \cellcolor{DeepBlue}\textcolor{white}{Q4\_K\_M} 
  & \cellcolor{DeepBlue}\textcolor{white}{Q5\_K\_M} 
  & \cellcolor{DeepBlue}\textcolor{white}{Q6\_K} 
  & \cellcolor{DeepBlue}\textcolor{white}{Q8\_0} 
  & \cellcolor{DeepBlue}\textcolor{white}{FP16} \\
& \textbf{PPL}
  & - & \makecell{6.7997\\[-0.4em]\textcolor{orange}{\tiny $\pm0.0433$}} & \makecell{6.2621\\[-0.4em]\textcolor{orange}{\tiny $\pm0.0392$}} & \makecell{6.0862\\[-0.4em]\textcolor{orange}{\tiny $\pm0.0380$}} & \makecell{6.0080\\[-0.4em]\textcolor{orange}{\tiny $\pm0.0374$}} & \makecell{5.9949\\[-0.4em]\textcolor{orange}{\tiny $\pm0.0373$}} & \makecell{5.9762\\[-0.4em]\textcolor{orange}{\tiny $\pm0.0371$}} & - \\
& \textbf{Tokens/s}
  & - & 139.78 & 137.34 & 136.93 & 191.87 & 196.46 & 143.49 & - \\
& \textbf{Param(GB)}
  & - & 12.0 & 15.0 & 20.0 & 22.0 & 26.0 & 33.0 & - \\
& \textbf{ARC-C}
  & - & 0.92 & 0.94 & 0.95 & 0.95 & 0.95 & 0.983 & - \\
& \textbf{ARC-E}
  & - & 1.0 & 1.0 & 1.0 & 1.0 & 1.0 & 1.0 & - \\
& \textbf{MMLU}
  & - & 0.723 & 0.814 & 0.824 & 0.824 & 0.826 & 0.824 & - \\
\midrule

%% ============ Qwen2.5-72B (补充) ============
\multirow{7}{*}{\textbf{Qwen2.5-72B}}
& \cellcolor{DeepBlue}\textbf{\textcolor{white}{Quant Scheme}} 
  & \cellcolor{DeepBlue}\textcolor{white}{IQ1\_M} 
  & \cellcolor{DeepBlue}\textcolor{white}{Q2\_K} 
  & \cellcolor{DeepBlue}\textcolor{white}{Q3\_K\_M} 
  & \cellcolor{DeepBlue}\textcolor{white}{Q4\_K\_M} 
  & \cellcolor{DeepBlue}\textcolor{white}{Q5\_K\_M} 
  & \cellcolor{DeepBlue}\textcolor{white}{Q6\_K} 
  & \cellcolor{DeepBlue}\textcolor{white}{Q8\_0} 
  & \cellcolor{DeepBlue}\textcolor{white}{-} \\
& \textbf{PPL}
  & \makecell{7.2507\\[-0.4em]\textcolor{orange}{\tiny $\pm0.0468$}} & \makecell{6.3354\\[-0.4em]\textcolor{orange}{\tiny $\pm0.0412$}} & \makecell{5.5831\\[-0.4em]\textcolor{orange}{\tiny $\pm0.0348$}} & \makecell{5.3012\\[-0.4em]\textcolor{orange}{\tiny $\pm0.0325$}} & \makecell{5.2964\\[-0.4em]\textcolor{orange}{\tiny $\pm0.0327$}} & \makecell{5.2720\\[-0.4em]\textcolor{orange}{\tiny $\pm0.0325$}} & \makecell{5.2644\\[-0.4em]\textcolor{orange}{\tiny $\pm0.0325$}} & - \\
& \textbf{Tokens/s}
  & 96.60 & 100.13 & 83.56 & 62.33 & 95.96 & 108.05 & 177.27 & - \\
& \textbf{Param(GB)}
  & 23.7 & 29.8 & 37.7 & 47.4 & 47.0 & 56.0 & 73.0 & - \\
& \textbf{ARC-C}
  & 0.95 & 0.94 & 0.97 & 0.97 & 0.97 & 0.97 & 0.97 & - \\
& \textbf{ARC-E}
  & 0.99 & 1.0 & 1.0 & 1.0 & 1.0 & 1.0 & 1.0 & - \\
& \textbf{MMLU}
  & 0.776 & 0.818 & 0.819 & 0.839 & 0.835 & 0.842 & 0.843 & - \\
\bottomrule
\label{tab:perplexity-results}
\end{tabular}
\end{table*}

In order to systematically evaluate the impact of model quantization on inference performance, we conducted comprehensive experiments on multiple models quantized via the GGUF framework. These experiments covered progressive quantization from 1-bit to 16-bit precision, focusing on well-performing yet moderately sized LLaMA-3.1 and Qwen-2.5 model families. Furthermore, given the remarkable performance recently exhibited by DeepSeek-R1, we also incorporated several DeepSeek-R1-distilled variants of LLaMA-3.1 and Qwen-2.5, such as DeepSeek-R1-Distill-Llama-8B and DeepSeek-R1-Distill-Qwen-14B, to enable detailed comparisons against the instruction-tuned Llama-3.1-8B-Instruct and Qwen2.5-7B-Instruct.

First, we used the WikiText-2 dataset to measure both the perplexity (PPL) degradation and token generation speed for each model under varying quantization bitwidths and strategies. As shown in Table~\ref{tab:perplexity-results}, our results provide a clear illustration of how quantization levels affect model performance.

\begin{figure}[!htbp]
    \centering
    \includegraphics[width=0.95\linewidth]{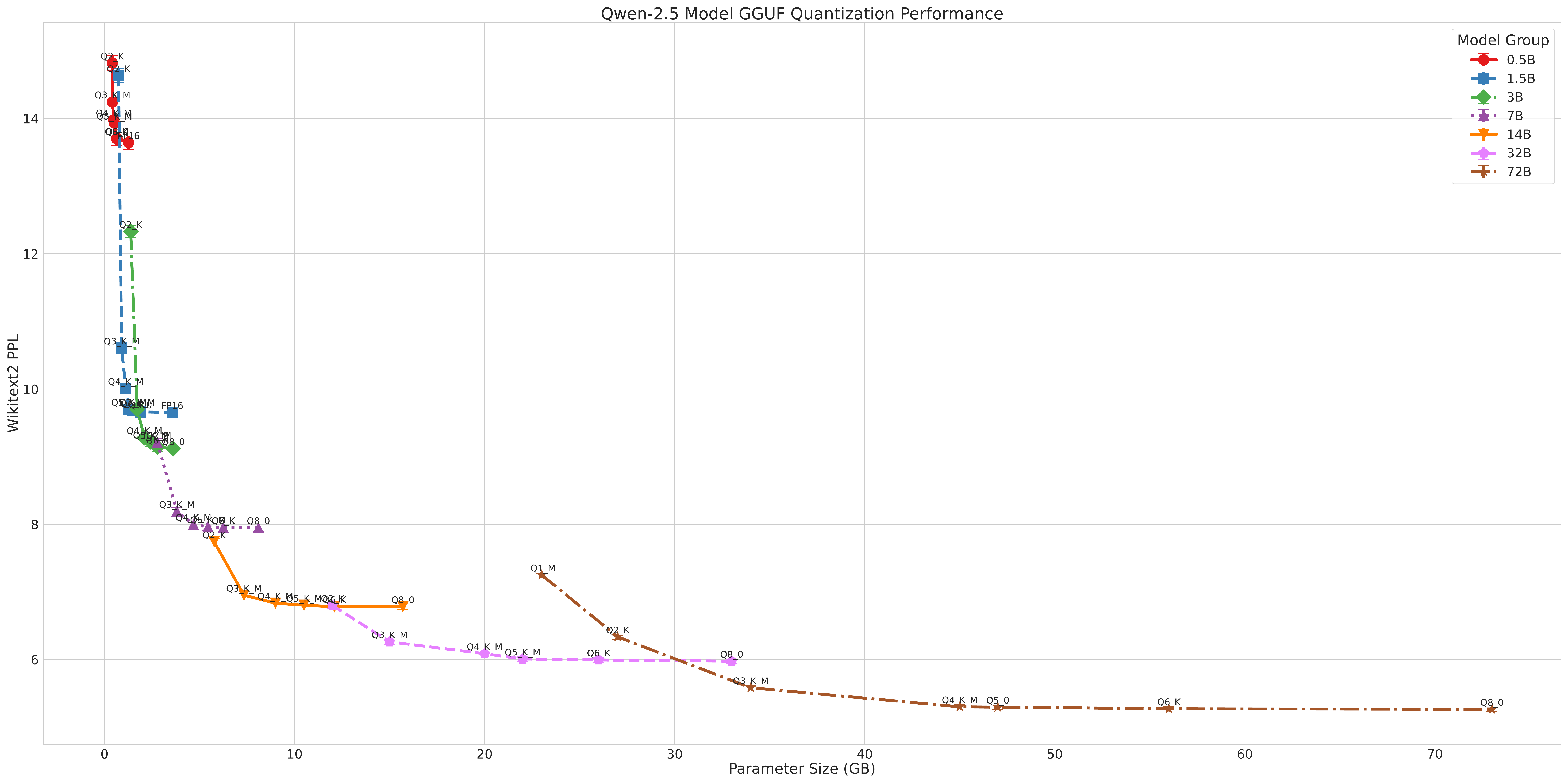}
    \caption{ The perplexity (PPL) performance of the Qwen-2.5 model on the Wikitext2 dataset under various GGUF quantization schemes at three different scales (0.5B, 1.5B, 3B, 7B, 14B, 32B and 72B).}
    \label{fig:quant_performance}
\end{figure}
\begin{figure}[!htbp]
    \centering
    \includegraphics[width=0.95\linewidth]{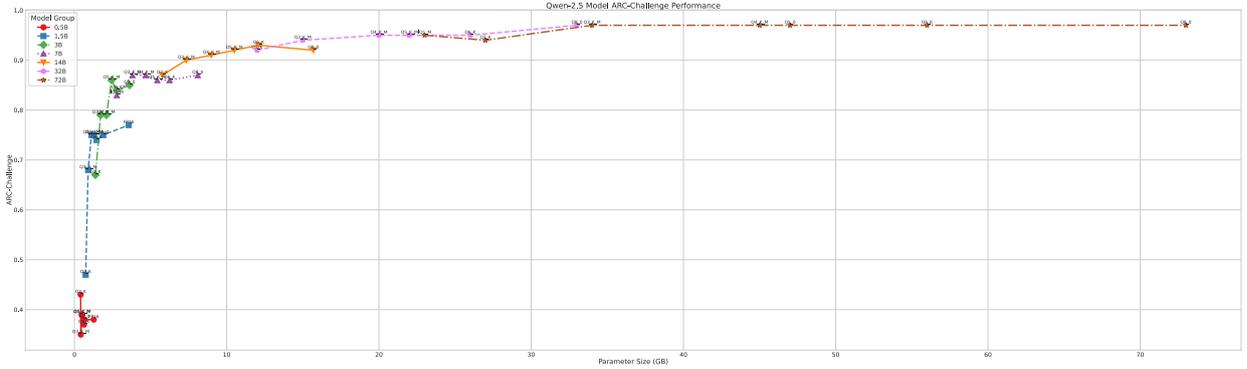}
    \caption{ The ARC-Challenge accuracy performance of the Qwen-2.5 model under various GGUF quantization schemes at three different scales (0.5B, 1.5B, 3B, 7B, 14B, 32B and 72B).}
    \label{fig:quant_performance1}
\end{figure}
\begin{figure}[!htbp]
    \centering
    \includegraphics[width=0.95\linewidth]{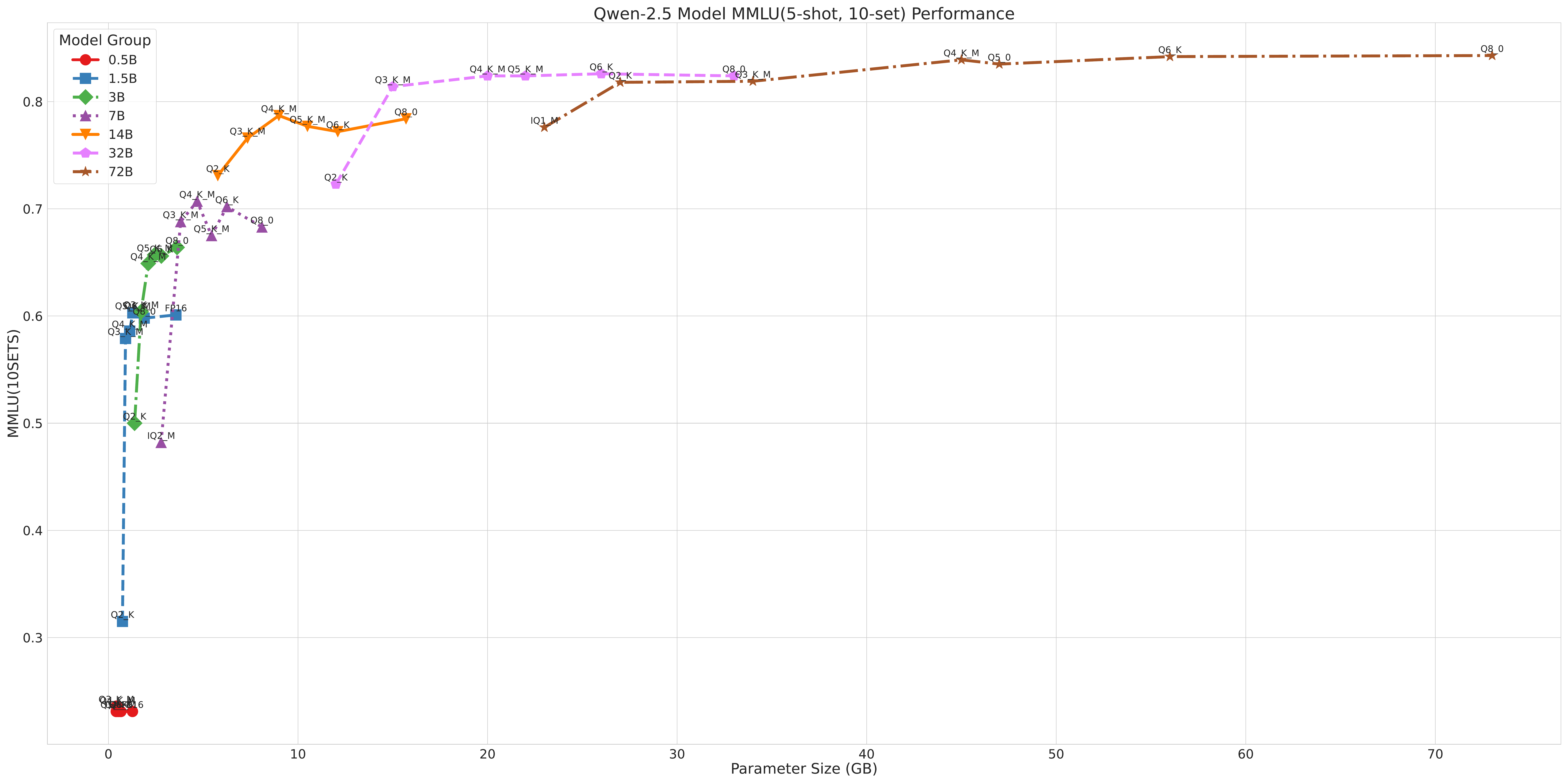}
    \caption{ The MMLU accuracy of the Qwen-2.5 model under various GGUF quantization schemes at three different scales (0.5B, 1.5B, 3B, 7B, 14B, 32B and 72B).}
    \label{fig:quant_performance2}
\end{figure}

\textbf{Efficiency of Compression Strategies:} Experimental findings reveal that 2-bit quantization can reduce model parameter sizes to merely one-fifth of their half-precision counterparts while preserving 91.4\% of baseline performance across all metrics. For instance, DeepSeek-R1-Distill-Llama-8B at Q2\_K achieves 89.7\% WikiText-2 performance (Fig \ref{fig:quant_performance}), 87.3\% ARC-C accuracy (Fig \ref{fig:quant_performance1}), and 88.2\% MMLU proficiency (Fig \ref{fig:quant_performance2}) compared to FP16, demonstrating superior compression efficiency over structured pruning.

\textbf{Threshold for Quantization Precision:} We observe a critical 3-bit threshold where performance degradation becomes non-linear. The Qwen2.5-14B model shows less than 3\% accuracy drop in ARC-C (0.90 vs 0.92) and MMLU (0.766 vs 0.787) at Q3\_K\_M, but suffers 5.8\% and 8.1\% respective drops at Q2\_K. This "safe compression zone" suggests 3-bit quantization optimally balances efficiency and capability preservation.

\textbf{Analysis of Quantization Efficiency:} Under equivalent experimental conditions, quantization significantly accelerates the inference speed. Specifically, 2-bit quantization (Q2\_K) achieves a one- to two-fold inference speedup compared with 8-bit quantization (Q8\_0), demonstrating the advantages of low-bitwidth quantization in memory-bound and latency-sensitive scenarios.

\textbf{Comparison of Instruction-Tuned Models and DeepSeek-R1-Distilled Models:} For the same parameter scale, the DeepSeek-R1-distilled LLaMA-3.1 and Qwen-2.5 models exhibit notably inferior PPL results on WikiText-2 compared with their instruction-tuned counterparts, implying that knowledge distillation may introduce certain semantic biases. Nevertheless, these distilled models generally show some benefit in inference speed, underscoring one of the key strengths of distillation strategies.

As an ablation study, we evaluated several widely used Post-Training Quantization (PTQ) methods on LLaMA2-7b, using the WikiText2 dataset and a maximum input length of 2K. The evaluation focused on PPL (Perplexity) to assess the model's accuracy retention after quantization (Table \ref{tab:quantization}). The results showed that most quantization methods maintain good accuracy, with a maximum precision loss of only 5\% at 4-bit quantization. In comparison to the three pruning methods tested earlier (structured, unstructured, and semi-structured pruning), the quantized models demonstrate a significant advantage in model robustness under the same compression rates, particularly in terms of reducing model degradation during compression.

\begin{table}[ht]
\centering
\small
\caption{Wikitext2-PPL results for various quantization methods with different bit configurations(2K).}
\label{tab:quantization}
\begin{tabular}{|l|c|c|c|c|}
\hline
Method & 2bits & 3bits & 4bits & 8bits \\
\hline
GPTQ & 1784.1625 & 7.5768 & 5.7459 & \underline{5.4739} \\
AWQ & - & \underline{6.2431} & \underline{5.6009} & - \\
GGUF & \textbf{5.8619} & \textbf{5.5463} & \textbf{5.4549} & \textbf{5.3976} \\
QLoRA\_NF4 & - & - & 5.6500 & - \\
QLoRA\_FP4 & - & - & 5.7700 & - \\
LLM.int8() & - & - & - & 5.5000 \\
SmothQuant w8a8 & - & - & - & 5.5934 \\
\hline
\end{tabular}
\end{table}

\subsection{Compression-Induced Model Hemorrhage: Quantization vs. Pruning}\label{sec:quantization_comp}
\begin{figure}[h]
    \centering
    \includegraphics[width=0.9\linewidth]{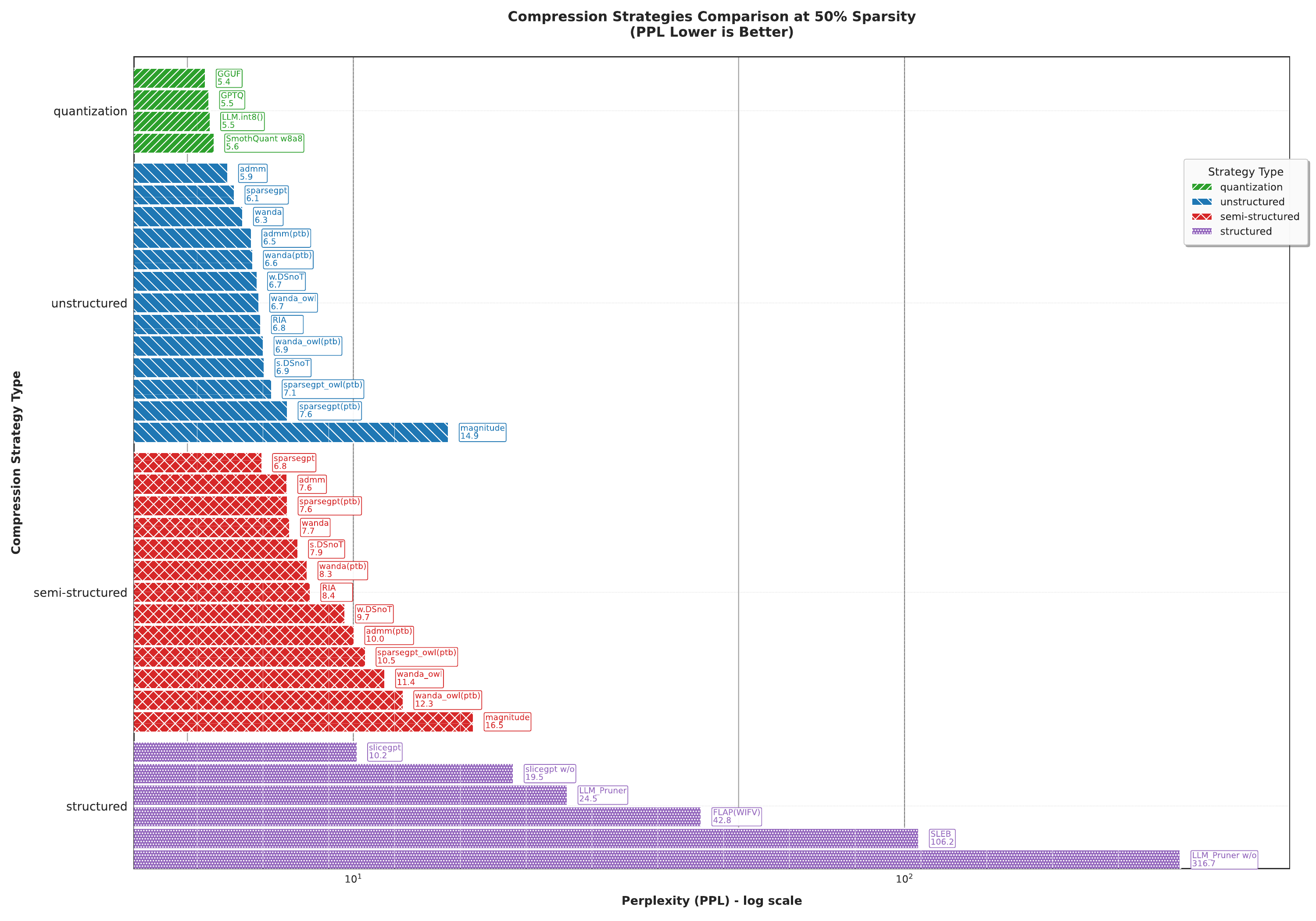}
    \caption{Comparison of perplexity (PPL) for quantization and pruning methods at 50\% compression rate on Wikitext2. Lower PPL indicates better performance. Quantization outperforms all pruning strategies, with unstructured pruning showing moderate robustness compared to semi-structured and structured approaches.}
    \label{fig:quant_compare}
\end{figure}

\begin{figure}[h]
    \centering
    \includegraphics[width=0.7\linewidth]{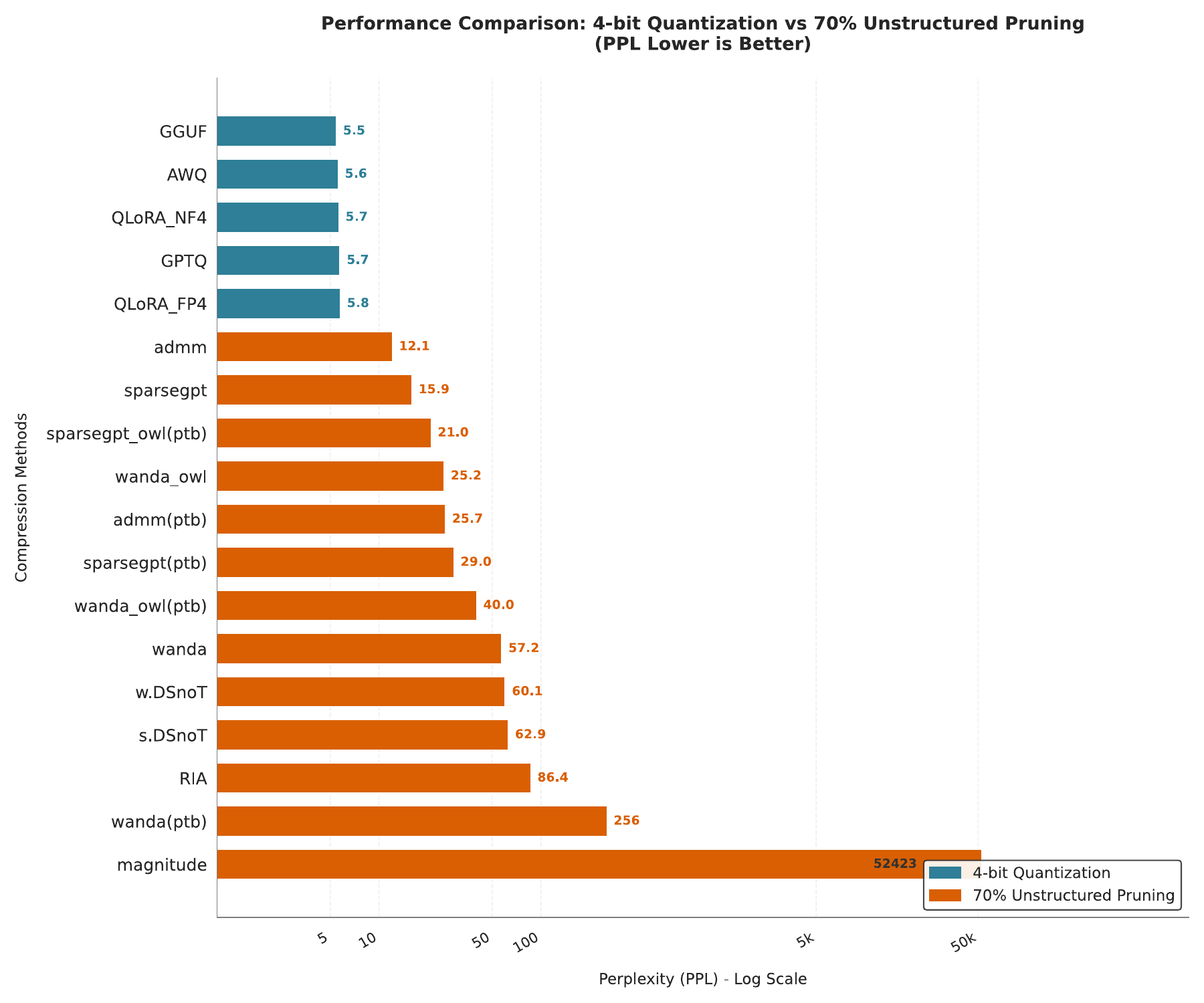}
    \caption{High-compression regime (70\% sparsity or 4-bit quantization) performance on Wikitext2. Quantization (e.g., GGUF: 5.5 PPL) demonstrates exceptional precision retention, while unstructured pruning methods (e.g., wanda: 57.2 PPL) suffer significant degradation, highlighting the challenges of maintaining accuracy under extreme compression.}
    \label{fig:quant_compare1}
\end{figure}

In this experiment, we compare various model compression methods, including both pruning and quantization techniques, under a 50\% sparsity setting (Fig \ref{fig:quant_compare}). The LLaMA2-7b model was tested on the WikiText2 dataset using Perplexity (PPL) as the performance metric. The results indicate that quantization outperforms pruning methods in terms of accuracy retention, with the ranking of performance being: Quantization > Unstructured Pruning > Semi-structured Pruning > Structured Pruning. Given that semi-structured and structured pruning methods demonstrated significantly lower robustness compared to unstructured pruning, we further compared quantization with unstructured pruning under higher compression rates (70\% sparsity with 4-bit quantization). The results show that quantization achieves outstanding accuracy retention, while many unstructured pruning methods begin to exhibit significant performance degradation (Fig \ref{fig:quant_compare1}).

\subsection{Sensitivity to Decoding Methods}\label{sec:decoding_lab}
We evaluate the performance of various decoding strategies on nine models (7B–9B) using both closed-ended tasks (MBPP, GSM8K) and open-ended tasks (Wikinews). Table~\ref{tab:performance} presents the results. Our experimental setup closely follows the approach outlined by Shi et al. in their analysis of decoding methods \cite{decoding_methods_llms}. The decoding methods used are summarized in Table~\ref{tab:decoding_methods}.

It is evident that the choice of decoding strategy has a notable impact on the model's performance. Among the deterministic methods, the Frustratingly Simple Decoding (FSD) and FSD-d methods stand out for their stability and strong performance, although they occasionally exhibit weaker results in open-ended text generation tasks. In contrast, the Greedy method, the fastest in terms of decoding speed, performs significantly worse than the other methods. Regarding stochastic methods, $\eta$ shows relatively poor performance, while temperature sampling (Temp) provides stable and excellent results.

The sensitivity of different models to decoding methods warrants further investigation. For larger and more powerful models, such as Yi-1.5-9B-Chat-16K and Gemma-2-9b-IT, the choice of decoding method seems to have a minimal impact across various tasks. However, for models like DeepSeek-R1-Distill-Llama-8B, which have undergone sample fine-tuning, the choice of decoding method is more sensitive in deterministic tasks. This could be due to its relatively weaker performance in mathematical reasoning and code generation tasks, or the potential capacity limitations of the distilled model, making it more sensitive to diversity control.

\definecolor{best}{RGB}{221, 160, 221} % 最优值颜色 (浅紫色)
\definecolor{worst}{RGB}{255, 204, 153} % 最差值颜色 (浅橙色)
\definecolor{secondbest}{RGB}{240, 200, 240} % 次优颜色 (淡紫色)
\definecolor{secondworst}{RGB}{255, 224, 178} % 次劣颜色 (淡橙色)

\renewcommand{\arraystretch}{1.3} % 行距调整
\begin{table*}[!htbp]
\centering
\caption{Performance of models under deterministic and stochastic decoding methods.}
\label{tab:performance}
\resizebox{\textwidth}{!}{
\begin{tabular}{ll|cccccc|cccccc}
\toprule
\multirow{2}{*}{\textbf{Model}} & \multirow{2}{*}{\textbf{Dataset}} & \multicolumn{6}{c|}{\textbf{Deterministic Methods}} & \multicolumn{6}{c}{\textbf{Stochastic Methods}} \\
%\cmidrule(lr){3-8} \cmidrule(lr){9-14}
 &  & \textbf{Greedy} & \textbf{BS} & \textbf{DBS} & \textbf{CS} & \textbf{FSD} & \textbf{FSD-d} & \textbf{Temp} & \textbf{Top-p} & \textbf{Top-k} & $\eta$ & \textbf{Miro} & \textbf{Typical} \\ 
\cline{1-14}
\midrule
\multirow{3}{*}{\textbf{Yi-1.5-9B-Chat-16K} \cite{Yi2024}} 
& MBPP & \cellcolor{worst}48.60 & 50.80 & 51.00 & \cellcolor{best}52.00 & 49.80 & 49.30 & 50.80& \cellcolor{secondworst}49.20 & 50.00 & 49.80 & 50.40 & \cellcolor{secondbest}51.60 \\
& GSM8K & \cellcolor{worst}64.21 & 65.50 & \cellcolor{secondbest}67.55 & 66.41 & 66.64 & \cellcolor{best}68.84 & 67.54 & 66.86 & \cellcolor{secondworst}64.90 & 65.20 & 66.94 & 67.40 \\
& Wikinews & 24.37 & \cellcolor{worst}23.52 & 24.38 & 24.40 & \cellcolor{best}25.10 & 24.30 & 24.55 & 24.45 & \cellcolor{secondbest}24.60 & 24.50 & \cellcolor{secondworst}24.20 & 24.35 \\
\cline{1-14}
\midrule
\multirow{3}{*}{\textbf{Gemma-2-9b-IT}} 
& MBPP & \cellcolor{worst}54.80 & 56.80 & 56.20 & 57.40 & \cellcolor{secondbest}57.60 & 56.60 & 56.00 & \cellcolor{best}57.80 & 56.20 & 57.00 & 56.60 & \cellcolor{secondworst}56.00 \\
& GSM8K & \cellcolor{worst}77.93 & \cellcolor{secondbest}79.15 & 78.62 & \cellcolor{best}79.30 & 78.89 & 79.15 & 78.86 & 78.64 & \cellcolor{secondworst}78.54 & 78.69 & 78.77 & 78.62 \\
& Wikinews & 25.91 & 25.98 & \cellcolor{secondworst}25.77 & 25.88 & 25.81 & 25.97 & \cellcolor{secondbest}26.12 & 25.99 & \cellcolor{best}26.13 & \cellcolor{worst}25.66 & 26.07 & 25.98 \\
\cline{1-14}
\midrule
\multirow{3}{*}{\textbf{Glm-4-9b-chat}} 
& MBPP & \cellcolor{worst}53.20 & \cellcolor{secondbest}57.20 & 55.40 & 55.80 & 56.40 & 56.40 & \cellcolor{best}60.00 & 54.60 & 57.00 & \cellcolor{secondworst}53.40 & 55.00 & 54.20 \\
& GSM8K & \cellcolor{worst}69.44 & 71.19 & \cellcolor{best}72.47 & 71.49 & 70.50 & \cellcolor{secondbest}72.25 & 68.76 & 70.76 & \cellcolor{secondworst}69.82 & 71.72 & 70.43 & 72.10 \\
& Wikinews & \cellcolor{secondbest}31.10 & \cellcolor{worst}30.40 & 30.68 & 30.66 & \cellcolor{best}31.27 & 30.87 & 30.89 & 30.64 & 31.00 & \cellcolor{secondworst}30.47 & 30.48 & 30.81 \\
\cline{1-14}
\midrule
\multirow{3}{*}{\textbf{Qwen2.5-7B-Instruct}} 
& MBPP & \cellcolor{worst}61.80 & \cellcolor{secondbest}66.00 & 63.40 & 65.00 & 63.80 & 65.00 & \cellcolor{best}66.40 & 64.40 & \cellcolor{secondworst}62.60 & \cellcolor{secondworst}62.60 & 63.40 & 64.00 \\
& GSM8K & \cellcolor{worst}77.33 & 78.85 & \cellcolor{secondbest}79.00 & 78.47 & \cellcolor{best}79.30 & 78.69 & 77.93 & 77.89 & \cellcolor{secondworst}77.71 & 78.62 & 78.40 & 78.85 \\
& Wikinews & \cellcolor{worst}30.31 & 30.52 & \cellcolor{secondbest}30.73 & 30.68 & 30.70 & 30.56 & 30.64 & 30.64 & \cellcolor{best}30.89 & 30.61 & \cellcolor{worst}30.31 & \cellcolor{secondworst}30.47 \\
\cline{1-14}
\midrule
\multirow{3}{*}{\textbf{DeepSeek-R1-Distill-Llama-8B}} 
& MBPP & \cellcolor{secondworst}6.80 & \cellcolor{worst}6.60 & 7.80 & 7.40 & 7.20 & \cellcolor{secondbest}8.20 & 8.00 & 7.60 & 7.60 & 7.00 & \cellcolor{secondbest}8.00 & 7.20 \\
& GSM8K & 16.15 & 16.91 & 17.13 & 16.45 & 17.44 & \cellcolor{secondbest}17.82 & \cellcolor{best}18.04 & 16.00 & 16.75 & \cellcolor{worst}15.24 & 16.45 & \cellcolor{secondworst}15.62 \\
& Wikinews & 19.66 & 19.70 & 19.75 & \cellcolor{secondworst}19.51 & \cellcolor{best}20.06 & 19.74 & 19.82 & 19.71 & 19.71 & \cellcolor{worst}19.32 & 19.79 & \cellcolor{secondbest}20.03 \\
\cline{1-14}
\midrule
\multirow{3}{*}{\textbf{DeepSeek-R1-Distill-Qwen-7B}} 
& MBPP & \cellcolor{best}4.20 & 3.20 & \cellcolor{secondworst}2.20 & 3.00 & \cellcolor{secondbest}3.60 & 3.40 & 2.60 & 2.80 & \cellcolor{worst}2.00 & 3.40 & 3.60 & 2.80 \\
& GSM8K & \cellcolor{worst}46.17 & 54.36 & 54.28 & 54.28 & 53.52 & \cellcolor{best}56.10 & \cellcolor{secondbest}54.96 & 53.90 & \cellcolor{secondworst}52.46 & 53.15 & 53.90 & \cellcolor{best}56.10 \\
& Wikinews & \cellcolor{secondworst}18.04 & 18.74 & 18.65 & \cellcolor{best}19.15 & \cellcolor{secondbest}19.09 & \cellcolor{worst}18.02 & 18.79 & 18.53 & 18.20 & 18.79 & 18.33 & 18.67 \\
\cline{1-14}
\midrule
\multirow{3}{*}{\textbf{DeepSeek-R1-Distill-Qwen-14B}} 
& MBPP & 3.40 & 3.60 & 3.40 & \cellcolor{best}4.40 & 3.60 & 3.60 & \cellcolor{secondworst}3.00 & 3.60 & 3.20 & 3.20 & \cellcolor{worst}2.80 & \cellcolor{secondbest}4.20 \\
& GSM8K & \cellcolor{worst}48.82 & 57.16 & 56.03 & 56.71 & 55.72 & \cellcolor{best}58.76 & 56.25 & 55.34 & \cellcolor{secondworst}55.12 & 57.62 & 56.18 & \cellcolor{secondbest}57.92 \\
& Wikinews & 25.24 & \cellcolor{secondworst}24.73 & 24.84 & 25.16 & \cellcolor{worst}24.69 & 25.34 & 24.80 & 25.33 & 25.43 & 24.96 & \cellcolor{secondbest}25.46 & \cellcolor{best}25.56 \\
\cline{1-14}
\midrule
\multirow{3}{*}{\textbf{Llama-3.1-8B-Instruct}} 
& MBPP & \cellcolor{worst}53.80 & 56.00 & 55.20 & 56.20 & 56.40 & \cellcolor{secondbest}57.20 & 56.60 & \cellcolor{best}58.80 & \cellcolor{worst}53.80 & \cellcolor{secondworst}54.20 & 55.20 & 55.40 \\
& GSM8K & \cellcolor{worst}72.93 & \cellcolor{secondworst}73.46 & 74.68 & 73.84 & 74.37 & \cellcolor{secondbest}74.83 & \cellcolor{best}77.18 & 73.46 & 73.69 & 74.60 & 74.07 & \cellcolor{worst}72.93 \\
& Wikinews & 15.80 & 16.20 & 16.22 & \cellcolor{best}16.61 & 15.65 & 15.98 & 15.72 & \cellcolor{secondworst}15.54 & 16.26 & \cellcolor{worst}15.17 & \cellcolor{secondbest}16.42 & 15.95 \\
\cline{1-14}
\midrule
\multirow{3}{*}{\textbf{Marco-o1-7B}} 
& MBPP & 14.20 & 14.80 & 14.60 & \cellcolor{best}16.20 & \cellcolor{secondbest}15.80 & 14.00 & 14.00 & 14.80 & 14.40 & \cellcolor{secondworst}13.80 & \cellcolor{worst}13.40 & 15.40 \\
& GSM8K & \cellcolor{worst}34.34 & 43.29 & 42.61 & \cellcolor{secondworst}41.32 & 42.53 & 43.21 & \cellcolor{best}46.10 & 42.00 & 42.00 & 43.06 & 41.62 & \cellcolor{secondbest}43.52 \\
& Wikinews & 34.14 & 34.08 & 34.04 & 34.01 & 34.05 & \cellcolor{secondbest}34.13 & 34.12 & 33.99 & \cellcolor{worst}33.89 & \cellcolor{secondworst}33.96 & \cellcolor{best}34.22 & 33.84 \\
\bottomrule
\end{tabular}}
\end{table*}

\begin{table*}[htbp]
\centering
\caption{Overview of Deterministic and Stochastic Decoding Methods in Large Language Models.}
\label{tab:decoding_methods}
\resizebox{\textwidth}{!}{
\begin{tabular}{|>{\centering\arraybackslash}m{4cm}|>{\centering\arraybackslash}m{5cm}|m{13.5cm}|}
\hline
\rowcolor{gray!20} \textbf{Category} & \textbf{Method} & \textbf{Description and Formula} \\
\hline
\multirow{25}{*}{\textbf{Deterministic Methods}} 
& Greedy Search & At each time step \( t \), selects the token with the highest probability:  
\[ y_t = \arg\max_{y \in V} P(y|x, y_{<t}) \]  
It can get stuck in local optima as it ignores global sequence scores. \\ \cline{2-3}

& Beam Search \cite{freitag2017beam} & Maintains the top \( k \) probable sequences at each step:  
\[ B_t = \arg\text{top}_k \left\{ \log P(y_{<t}, y | x) \right\} \]  
\( k \) is the beam width, and \( k=4,8 \) in experiments. \\ \cline{2-3}

& Diverse Beam Search \cite{vijayakumar2018diverse} & Improves diversity by grouping \( k \) sequences into \( G \) groups:  
\[
\text{score}(y_{<t}, y) = \log P(y_{<t}, y | x) - \lambda \sum_{g' < g} \Delta((y_{<t}, y), B^g_t)
\]
\( \lambda=1 \) controls the diversity penalty. \\ \cline{2-3}

& Contrastive Search \cite{su2022contrastive} & Adds a penalty to tokens producing similar hidden states:  
\[
y_t = \arg\max_{y \in V_k} \left[ (1-\alpha)P(y|x, y_{<t}) - \alpha \max s(h_y, h_v) \right]
\]
\( \alpha \in [0.1, 0.6] \) balances similarity penalties. \\ \cline{2-3}

& Contrastive Decoding \cite{contrastive_decoding} & Penalizes undesired attributes from an "amateur" model:  
\[
\text{score} = (1 + \beta)u_y - \beta v_y, \quad \beta \in [0.1, 0.9]
\] \\ \cline{2-3}

& Frustratingly Simple Decoding (FSD) \cite{yang2023fsd} & Balances logits between LM and anti-LM:  
\[
\text{FSD}(y|x, y_{<t}) = (1-\alpha)P_\theta(y|x, y_{<t}) - \alpha P_\omega(y|x, y_{<t})
\]
\( \alpha \in [0.1, 0.6] \) adjusts balance. \\ \cline{2-3}

& DoLa \cite{dola} & Contrasts logits from the last layer and a premature layer using Jensen-Shannon Divergence (JSD):  
Premature layer is dynamically selected to maximize JSD. \\ \cline{2-3}

\hline

\multirow{23}{*}{\textbf{Stochastic Methods}}

% ------ 现有方法：Temperature / Top-k / Top-p / Typical / Top-η / Mirostat ------
& Temperature Sampling & Adjusts randomness by scaling logits using \( \tau \):  
\[
P(y|x, y_{<t}) = \frac{\exp(u_y / \tau)}{\sum_j \exp(u_j / \tau)}, \quad \tau \in [0.1, 0.9]
\] \\ \cline{2-3}

& Top-k Sampling \cite{hierarchical_story_generation} & Selects from the top-\( k \) tokens to avoid unreliable tail probabilities:  
\( k \in [5, 10, 20, 50, 100] \). \\ \cline{2-3}

& Top-p Sampling \cite{holtzman2020curious}& Chooses the minimal set \( V_p \) of tokens that sum to \( p \):  
\[
\sum_{y \in V_p} P(y|x, y_{<t}) \geq p, \quad p \in [0.8, 1.0]
\] \\ \cline{2-3}

& Typical Sampling \cite{meister2023typical} & Minimizes entropy-probability difference:  
\[
\min_{V_c} \sum_{y \in V_c} \left| H(Y_t|x, y_{<t}) + \log P(y|x, y_{<t}) \right|
\]
\( p \in [0.2, 0.95] \). \\ \cline{2-3}

& Top-\(\eta\) Sampling \cite{truncation_sampling} & Truncates words below an entropy-based threshold:  
\[
V_c = \left\{ y \in V : P(y|x, y_{<t}) \geq \sqrt{\eta} \exp(-h_\theta(x, y_{<t})) \right\}, \quad \eta \in [0.0003, 0.004]
\] \\ \cline{2-3}

& Mirostat Sampling \cite{basu2021mirostat} & 
Adaptively adjusts truncation size \( k_t \) to target perplexity \( \tau \):
\[
\begin{aligned}
\epsilon_t &= \tau - \log \hat{P}(y_t \mid y_{<t}), \\
k_{t+1} &= k_t + \alpha \,\epsilon_t
\end{aligned}
\]
\( \tau \in [2.5, 5] \). \\ \cline{2-3}

\hline 

\multirow{7}{*}{\textbf{Hybrid Methods}} 

& Stochastic Beam Search \cite{gumbel_topk} & Hybrid of beam search and sampling: 
\[ P(B_t) \propto \exp(\lambda \log P(y_{<t},y|x)),\ \lambda \in [0.5,2.0] \]
Uses Gumbel-Top-k trick for differentiability. \\ 
\cline{2-3}

& Lookahead Decoding \cite{lookahead_decoding} & Interleaves speculative drafting (greedy) and verification (sampling): 
\[ \text{AcceptRate} = \frac{\exp(P(y_d))}{\exp(P(y_s))} \]
Accelerates decoding while maintaining diversity. \\ 
\hline 

\end{tabular}
}
\end{table*}




\end{appendices}



\end{document}